\documentclass{bmvc2k}
\pdfoutput=1

\usepackage[utf8]{inputenc}
\usepackage[T1]{fontenc}
\usepackage{textcomp}
\usepackage[toc,page]{appendix}

\title{360\!° Optical Flow using Tangent Images}

\addauthor{Mingze Yuan}{https://yuanmingze.github.io/}{1}
\addauthor{Christian Richardt}{https://richardt.name}{1}

\addinstitution{
 University of Bath\\
 Bath, UK
}

\runninghead{Yuan and Richardt}{360° Optical Flow using Tangent Images}

\usepackage{xcolor}
\newcommand{\new}[1]{#1}
\newcommand{\footurl}[2]{\href{#1}{#2}\footnote{\url{#1}}}

\usepackage{microtype}
\usepackage{enumitem}
\usepackage{booktabs}
\usepackage{multirow}
\usepackage{tikz,siunitx}%
\usepackage{grffile}  %
\usepackage{doi}
\usepackage[nameinlink,capitalise,noabbrev]{cleveref}
\crefformat{equation}{#2Equation~#1#3}
\begin{document}

\maketitle

\begin{abstract}
\noindent
Omnidirectional 360° images have found many promising and exciting applications in computer vision, robotics and other fields, thanks to their increasing affordability, portability and their 360° field of view.
The most common format for storing, processing and visualising 360° images is equirectangular projection (ERP).
However, the distortion introduced by the nonlinear mapping from 360° images to ERP images is still a barrier that holds back ERP images from being used as easily as conventional perspective images.
This is especially relevant when estimating 360° optical flow, as the distortions need to be mitigated appropriately.
In this paper, we propose a 360° optical flow method based on tangent images.
Our method leverages gnomonic projection to locally convert ERP images to perspective images, and uniformly samples the ERP image by projection to a cubemap and regular icosahedron faces, to incrementally refine the estimated 360° flow fields even in the presence of large rotations.
Our experiments demonstrate the benefits of our proposed method both quantitatively and qualitatively.
\end{abstract}

\section{Introduction}
\label{sec:intro}

Commercial 360° video cameras have recently grown in popularity.
Companies such as GoPro, Insta360, Ricoh, and many others, are now offering affordable consumer 360° video cameras.
At the same time, support for 360° videos has been added across the content creation and consumption pipeline, including video editing software, such as DaVinci Resolve and Adobe Premiere Pro,
video distribution channels such as YouTube, Facebook and Vimeo,
video players like VLC, and virtually all current VR headsets.

Optical flow is a crucial component for processing and editing 360° images and videos as it provides correspondences over time.
Among other things, this enables 360° video stabilisation \cite{Kopf2016}, depth estimation \cite{ZioulKZAD2019}, and novel-view synthesis \cite{BerteYLR2020}.
However, the most common format of 360° images, equirectangular projection (ERP), suffers from severe distortions when mapping a spherical 360° image to the 2D image plane.
This distortion is not considered by most existing optical flow algorithms,
which may lead to flow estimation errors.

Previous work has adapted existing CNN-based techniques designed for perspective images to ERP images by transforming convolution kernels to match the projection \cite{CoorsCG2018, SuG2019, TatenNT2018}.
To overcome the limited spatial resolution supported by these methods, other work proposed extracting, processing and recombining tangent images \cite{EderSLF2020, LuoZSX2019, ZhangLSC2019, LeeJYJY2019, WangHCLYSCS2018, WangYSCT2020}, which supports both CNN-based and traditional methods.
Moreover, these tangent images are based on \emph{gnomonic} projection, which is a common tool used in geography to map a sphere to a plane, as the great circle on the sphere maps to a straight line in the tangent image~\cite{EderSLF2020}.
This principle has been applied to tasks such as monocular depth estimation \cite{WangHCLYSCS2018, WangYSCT2020}, semantic segmentation \cite{EderSLF2020, ZhangLSC2019}, and classification \cite{EderSLF2020, LuoZSX2019, LeeJYJY2019}, but not yet to optical flow.
Our approach builds on this insight to estimate accurate 360° optical flow for high-resolution ERP images by sampling tangent images uniformly based on regular polyhedra (cube and icosahedron).
\looseness-1

This paper makes the following technical contributions:
\begin{enumerate}[nosep]
\item 
We propose a 360° optical flow method based on gnomonic projection to overcome the distortions induced by the equirectangular projection of 360° images.

\item
We integrate a global warping method to align large rotations between 360° images.

\item
We introduce a tangent image optical flow blending method that can dramatically reduce discontinuities at face boundaries.

\item 
We refine our flow estimates by increasing the sampling density of tangent images from an initial cubemap to the final icosahedron-based uniform sampling of tangent images.
\end{enumerate}

\section{Related Work}

\textbf{Optical Flow}
has been a fundamental computer vision technique for decades, as it estimates dense correspondences between two input images \cite{BakerSLRBS2011,MenzeHG2018}.
These correspondences can be used to identify object or camera ego-motion, and optical flow thus finds many applications in robotics, scene understanding, geometry reconstruction etc.
Traditional optical flow methods are formulated using energy minimization based on photoconsistency and regularisation with a smoothness term \cite{KroegTDV2016, HornS1981, LucasK1981, BlackA1991}.
\citet{BroxBPW2004} proposed a warping-based method that refines optical flow estimations in a coarse-to-fine fashion.
\citet{SunRB2014} uses a median filter post-processing step to produce sharper object boundaries in the optical flow fields.

In recent years, learning-based methods have defined a new state of the art in optical flow estimation.
The first methods were based on supervised learning, generally on synthetic training data.
FlowNet \cite{DosovFIHHGSCB2015} lifts the correlation operation into feature space, and uses a multi-scale architecture to effectively predict optical flow from two input images.
FlowNet2 \cite{IlgMSKDB2017} introduces image warping between multiple cascaded FlowNets to increase the accuracy of large pixel motion.
PWC-Net \cite{SunYLK2020} incorporates the best practices of traditional optical flow methods into a neural network: pyramid processing, image warping, and cost volume processing.
RAFT \cite{TeedD2020a} employs a recurrent network to iteratively estimate optical flow from the 4D correlation volume between all pairs of pixels. %
\citet{AleotPM2021} propose a data generation method using monocular depth estimation to synthesise a second view from a single input image.
This enables self-supervised training of existing optical flow methods.

Virtually all optical flow methods focus on perspective input images, although spherical images have recently received more attention.
To this end, recent approaches by \citet{ArtizZAD2020} and \citet{BhandZY2020}
transfer the architecture and weights of pre-trained networks \cite{DosovFIHHGSCB2015,HuiTC2018} to spherical images in the equirectangular projection (ERP) format.

\paragraph{Panoramic Image Processing.}

Panoramic (aka 360° or \emph{spherical}) images have been used in a wide variety of application areas, including
depth estimation \cite{ImHRJCK2016, JiangSZDH2021, LaiXLL2019, SunSC2021, WangHCLYSCS2018, WangSTCS2020, WangYSCT2020, ZioulKZD2018, ZioulKZAD2019},
room layout estimation \cite{WangYSCT2021, Tran2021, EderMG2019, FernaFPDCG2020, JinXZZTXYG2020, SunSC2021, ZengKG2020},
semantic segmentation \cite{LeeJYJY2019, SunSC2021, YangZRHS2021, ZhangLSC2019},
novel-view synthesis \cite{BerteYLR2020, HuangCCJ2017, MatzeCEKS2017, XuZXTG2021} and so on.
The key problem when working with spherical images is to project them onto a regular 2D pixel grid for easy processing, as any projection introduces some kind of distortion – similar to maps of the world.
\looseness-1 %

Spherical images are sometimes projected to a cubemap,
i.e. the six sides of a cube, potentially with overlap between sides \cite{ChengCDWLS2018, WangHCLYSCS2018, WangYSCT2020}.
However, the resulting perspective images have large angles of view of $\geq$90°, which introduces significant perspective distortion.
A smaller field of view can be achieved using icosahedron projection, producing 20 triangular faces, or in fact any subdivision of the icosahedron \citep{LuoZSX2019, ZhangLSC2019, LeeJYJY2019}.
In the limit, perspective tangent images can be obtained anywhere on the sphere using gnomonic projection \citep{CoorsCG2018, EderSLF2020}.
To overcome the significant distortions near the poles at the top/bottom image edges in ERP images, distortion-aware convolutions adapt kernels using gnomonic projection \cite{CoorsCG2018, SuG2019, TatenNT2018}, which enable training on perspective images and testing on ERP images.

Our method combines ERP, cubemap and icosahedron projections to incrementally estimate global rotation and to refine spherical optical flow estimates to achieve higher accuracy.

\section{Approach}
\label{sec:approach}

\begin{figure}%
	\centering
	\includegraphics[width=0.95\linewidth]{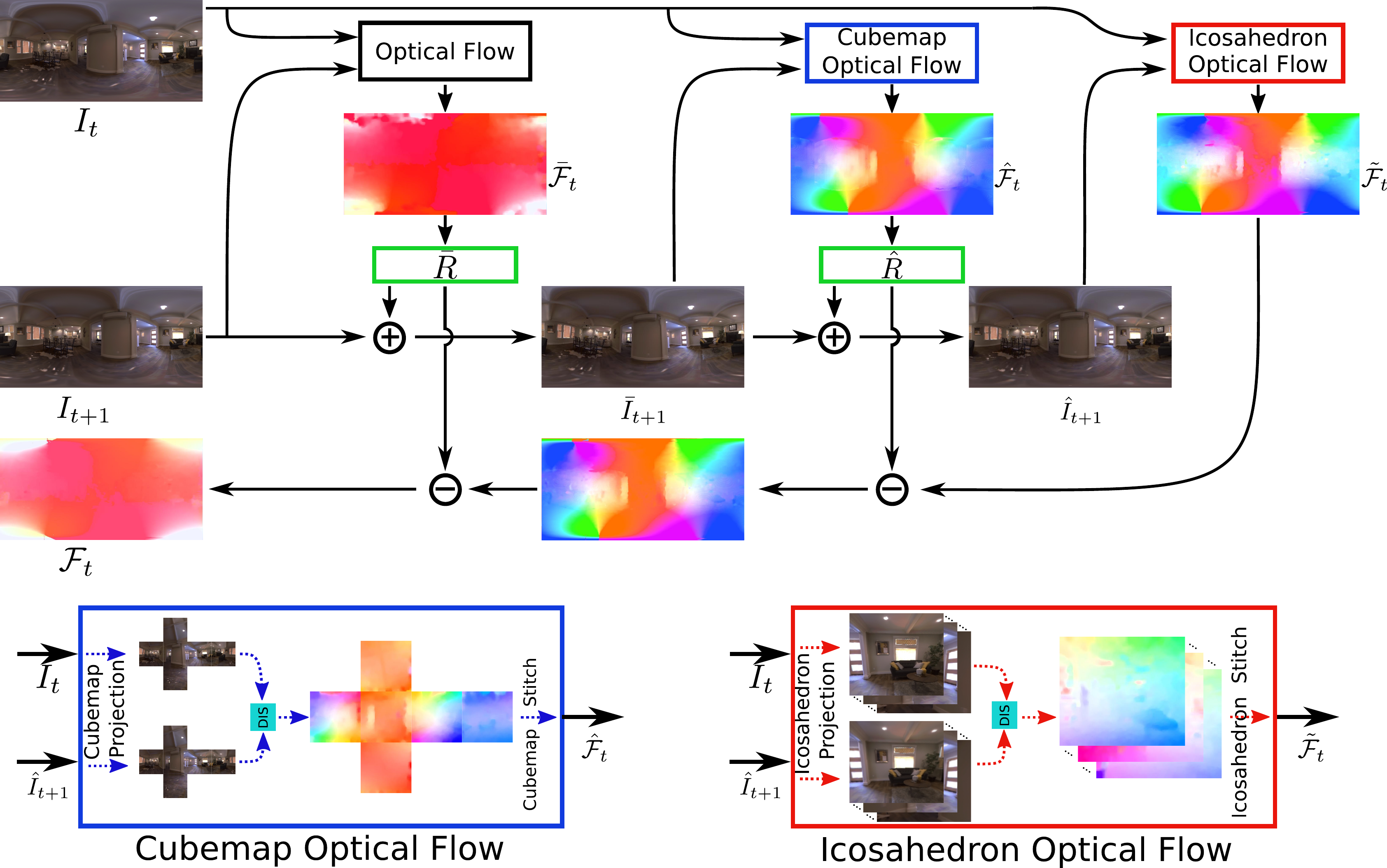}
	\caption{Method overview.
		The input is a pair of equirectangular images, $I_t$ and $I_{t+1}$, and the output is the optical flow $\mathcal{F}_t$.
		The operators `$\oplus$' and `$\ominus$' rotate images forward and optical flow fields backward, respectively, using image warping (see \cref{sec:approach:warping}).
		The green box estimates the optimal rotation from a 360° optical flow field.
	}
	\label{fig:approach:pipeline}
\end{figure}

Our method comprises a global rotation warping and tangent-image-based panoramic optical flow method to mitigate the distortions in equirectangular image and the large displacements of pixels.
At the same time, our high-level approach generalises to any off-the-shelf perspective optical flow method, and thus will benefit from future improvements in optical flow methods.

As shown in \cref{fig:approach:pipeline}, our method comprises three steps:
(1)~We estimate the optical flow $\bar{\mathcal{F}}_t$ from ERP image $I_{t}$ to ${I_{t+1}}$, followed by rotating $I_{t+1}$ towards to $I_{t}$ based on the rotation estimate $\bar{R}$ from the flow $\bar{\mathcal{F}}_t$, which generates the warped ERP image ${\bar{I}}_{t+1}$ (see \cref{sec:approach:warping}).
(2)~We estimate the panoramic optical flow ${\hat{\mathcal{F}}}_t$ from $I_{t}$ to ${\bar{I}}_{t+1}$ based on cubemap projection optical flow (see \cref{sec:approach:projstit}), and then rotate the image ${\bar{I}}_{t+1}$ according to the rotation $\hat{R}$ estimated from the cubemap flow ${\hat{\mathcal{F}}}_t$ to generate ${\hat{I}}_{t+1}$.
(3)~We estimate the fine-scale flow $\tilde{\mathcal{F}}_t$ from $I_{t}$ to ${\hat{I}}_{t+1}$ using icosahedron projection optical flow (see \cref{sec:approach:warping}), then backward-rotate the fine-scale flow $\tilde{\mathcal{F}}_t$ by rotations $\bar{R}$ and $\hat{R}$ to generate the final 360° optical flow $\mathcal{F}_t$.
\looseness-1

Compared to perspective images, equirectangular images are continuous in all direction.
In \cref{sec:approach:definition}, we analyse and define the panoramic optical flow.
To solve the equirectangular image distortion, especially at the top and bottom, our method employs gnomonic projection to project the equirectangular image to tangent image~\cite{EderSLF2020}.
We use cubemap and regular icosahedron faces to uniformly sample the equirectangular image to solve this problem (see \cref{sec:approach:projstit}).
However, compared to a panoramic image, a tangent image has a smaller field of view, not larger than 180°, and just 90° for cubemap projection (without padding).
The tangent image optical flow operates on a pair of corresponding  tangent images from the same ERP image area,
which can fail when an object is only visible in one of the tangent images.
To compensate for the tangent images' narrow field of view, the global rotation operation is used on the target image $I_{t+1}$ to pre-align it with the source image $I_t$ (\cref{sec:approach:warping}).

\subsection{Definition of 360° Optical Flow}
\label{sec:approach:definition}

\begin{figure}[hbt!]
	\centering
	\subfigure[Source Image]{\includegraphics[width=0.26\linewidth]{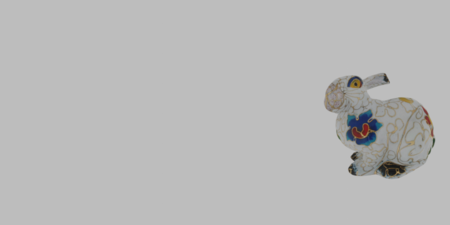}}
	\subfigure[Target Image]{\includegraphics[width=0.26\linewidth]{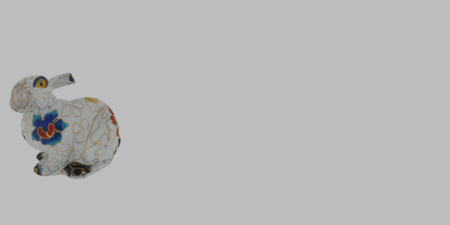}} 
	\subfigure[Perspective Optical Flow]{\includegraphics[width=0.26\linewidth]{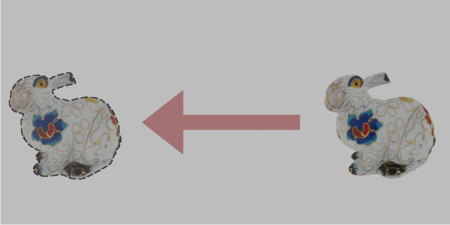}}
	\subfigure[Wrap-Around]{\includegraphics[width=0.19\linewidth]{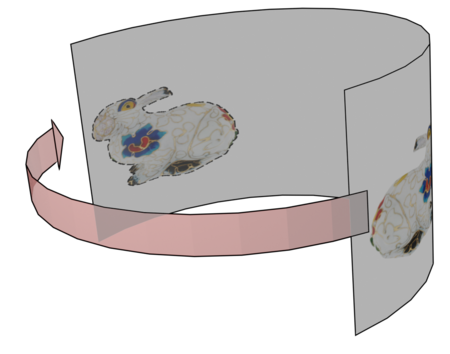}}
	\caption{\label{fig:app:warparound}%
		360° optical flow illustrated:
		A bunny moves from (a) to (b) within an ERP image.
		(c) Perspective optical flow methods estimate the bunny's motion from the right to the left via the front.
		(d) However, the shortest path for the bunny to move is along the back wrap-around.
	}
\end{figure}

Spherical image coordinates are continuous in any direction on the image, e.g. pixel locations overflowing the image width will wrap around to the other side of the image.
If panoramic optical flow follows an object, a pixel's motion vector can fall outside the image boundary.
This is one reason that perspective optical flow methods cannot track pixels moving outside the equirectangular image boundary, because they do not support the horizontal coordinate wrap-around, as illustrated in \cref{fig:app:warparound}.
However, this introduces an ambiguity into 360° optical flow estimation, 
as there is more than one path from the source point to the target point along a great-circle on the sphere:
usually there is one shorter and one longer path.\footnote{One could also travel along the great circle a few more times, but these paths are getting longer and longer.}
To uniquely define 360° optical flow in the equirectangular image format, we define the optical flow to follow the shortest path from source to target along the great circle between them.
This naturally limits the maximum flow magnitude to $\leq$180°.

\subsection{Projection \& Stitching}
\label{sec:approach:projstit}

To mitigate the distortions in ERP images, we locally undistort the ERP image to a perspective tangent image using gnomonic projection \cite{EderSLF2020}.
This way, we can apply any off-the-shelf perspective optical flow method directly on pairs of tangent images, first using a cubemap projection and later using an icosahedron projection to refine the 360° flow estimates.
In both the cubemap and icosahedron cases, we proceed as follows:
we first project the ERP image with gnomonic projection to generate 6 (cubemap) or 20 (icosahedron) perspective images that are tangent to the unit sphere;
we then use an optical flow method for perspective images to estimate optical flow for pairs of corresponding tangent images; and
finally, we stitch the optical flow fields for all tangent images back together in the ERP format.
Our approach in principle supports any perspective optical flow method, and results are therefore expected to improve as improved optical flow methods become available in future.

\begin{figure}[hbt!]
	\begin{center}
		\subfigure[Gnomonic Projection]{\includegraphics[width=0.57\linewidth]{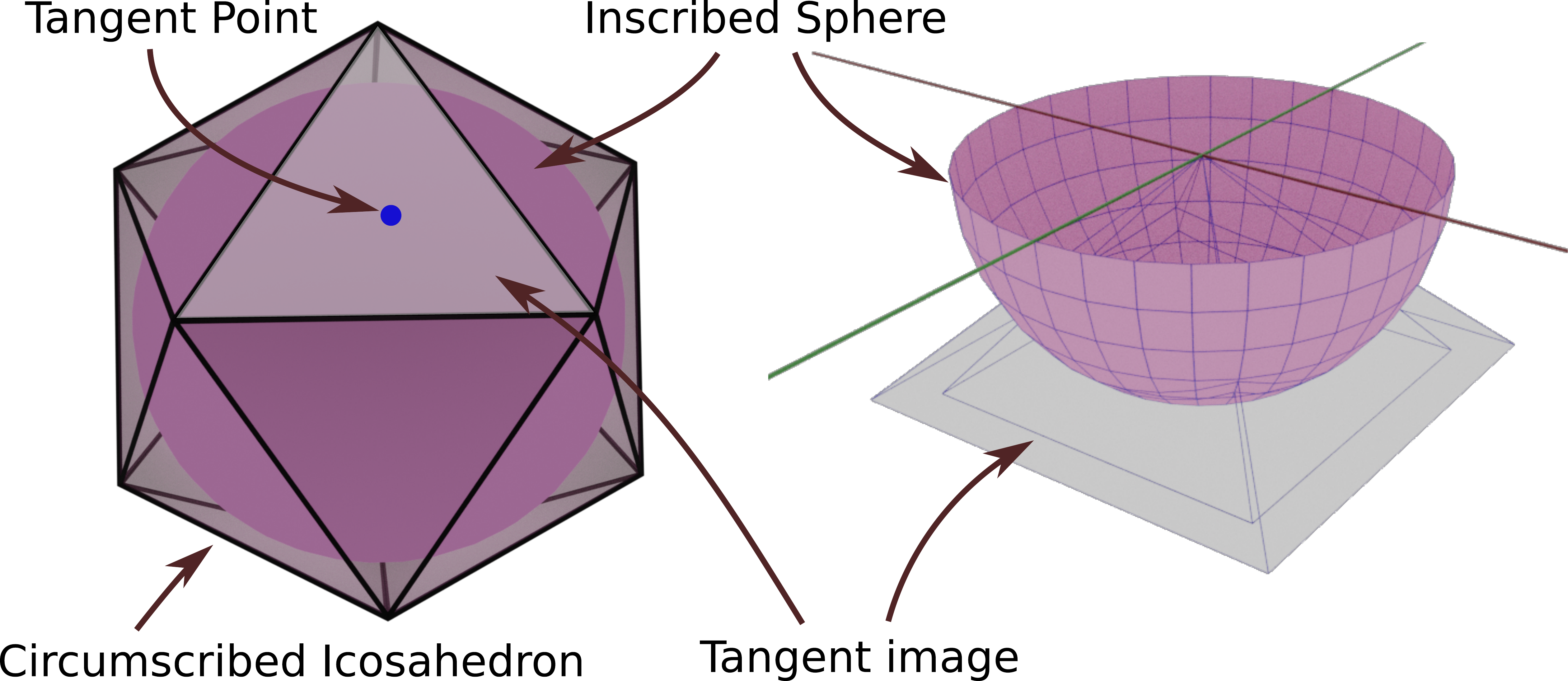}}
		\subfigure[Target Image Padding]{\includegraphics[width=0.35\linewidth]{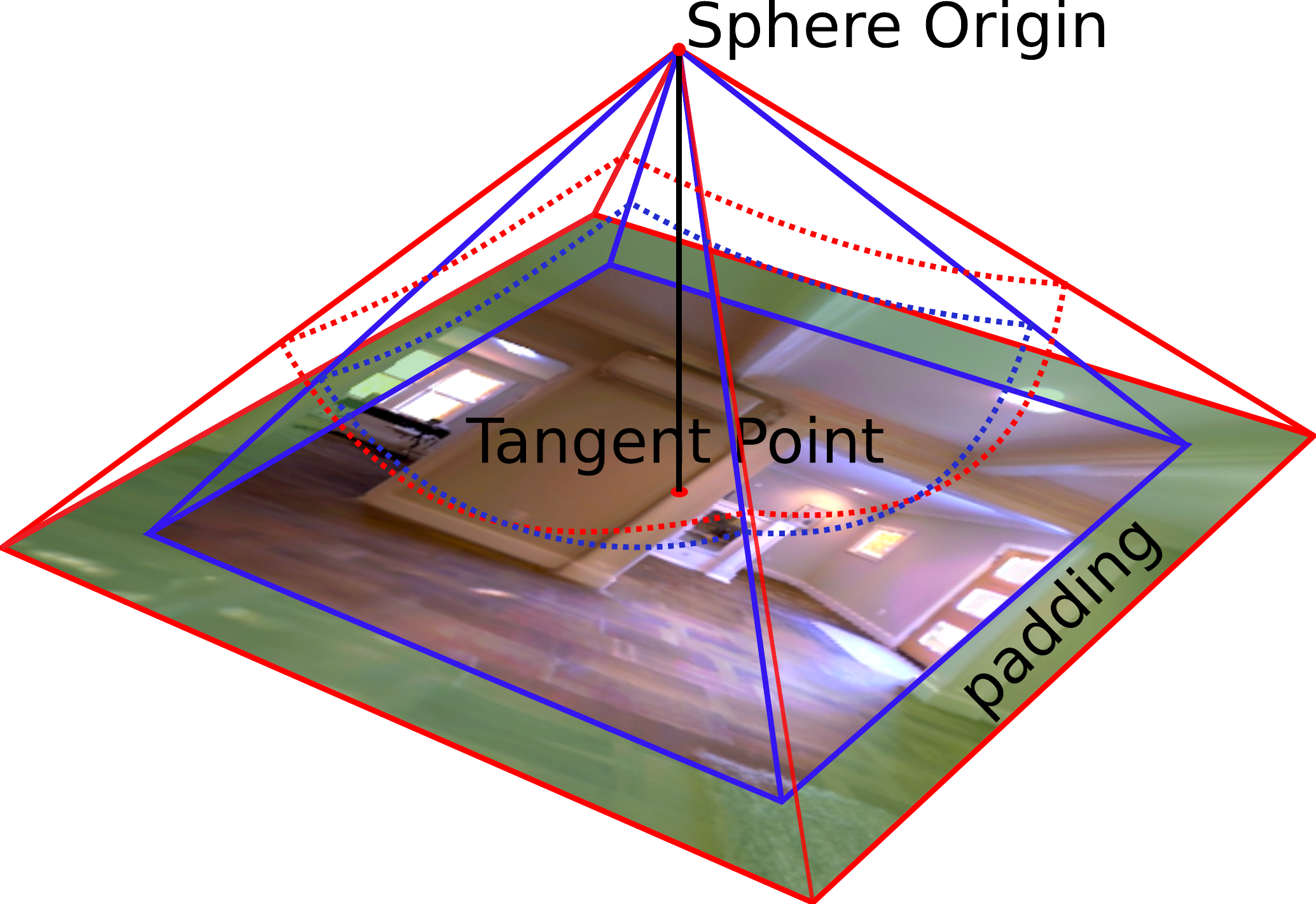}} 
	\end{center}
	\caption{\label{fig:approach:projection}%
		Tangent image projection and padding.
		(a)~Gnomonic projection maps points on the surface of a sphere from the sphere's centre to a point on the tangent plane.
		(b)~We add padding to the tangent images to extend their area and ensure overlap between tangent images.
	}
\end{figure}

\vspace{-1em}
\paragraph{Gnomonic Projection.}
Like \citet{ZhaoYZLBT2020}, we use the gnomonic projection to produce tangent images.
These images are perspective images with the same camera centre as the ERP image, but their principal axis intersects the unit sphere at the tangent point with a circumscribed cube or icosahedron.
Depending on the choice and density of tangent points, tangent images cover different spherical surface areas, which determines the field of view (FoV) of the tangent images.
For estimating multi-scale motion and to uniformly sample the sphere surface, we select tangent points according to a regular cubemap and a regular icosahedron.
These are the most common convex regular polyhedra with 6 and 20 faces, respectively, and this results in different FoV with cubemap tangent images having a considerably wider FoV than icosahedron tangent images.
Therefore, cubemaps can estimate a larger motion vectors from the tangent images, although their angular resolution is coarser than the icosahedron's tangent images for the same spatial tangent image resolution.
To increase the FoV of tangent images and improve the continuity of optical flow at the boundary between tangent images, we add image padding to extend the area of tangent images, as illustrated in \cref{fig:approach:projection}.

\paragraph{Tangent Optical Flow Blending.}
As each tangent image's optical flow is estimated independently, the estimates might not be consistent in the overlap areas of adjacent tangent images.
When stitching the tangent flow fields in the equirectangular format, we thus compute a per-pixel blending weight $\omega_{i}$ for each tangent image $I_{t,i}$ to smoothly blend its optical flows $\mathcal{F}_{t,i}$ in the overlap areas.
The $\omega_{i}$ computes the colour difference between the source tangent image $I_{t,i}$ and the backwards-warped target image, to reduce the weight of badly estimated tangent images' optical flow $\mathcal{F}_{t,i}$.
The $\omega_i$ is estimated by 
$e^{-| I_{t,i} - Warp(I_{t+1,i}, \mathcal{F}_{t,i})|}$, the $Warp(\cdot , \cdot)$ is the backwards warping operation, and $I_{t+1,i}$ is the corresponding tangent image of $I_{t,i}$ at time ${t+1}$.
The $\mathcal{F}_t$ blended with the weighted tangent image optical flows,  $\frac{\sum_i{\mathcal{F}_{t,i} \cdot \omega_{i}}}{\sum_i{\omega_i}}$.

\subsection{Global Rotation Warping}
\label{sec:approach:warping}

Inspired by warping-based optical flow~\cite{BroxBPW2004}, we estimate the global rotation to pre-align the ERP image $I_{t+1}$ to $I_{t}$.
This helps reduce the range of pixel motion to a level that is more suitable for the smaller FoV of tangent images.
We compute the global rotation $\bar{R}$ from the computed optical flow field as follows.
First, we convert the coordinates of the start and end points of all optical flow vectors to spherical coordinates with unit length.
Then, we solve for the optimal rotation between the start points and the corresponding end points using least-squares fitting, which can be solved in closed form by singular value decomposition \cite{ArunHB1987,SorkiR2017}.

\vspace{-1em}
\paragraph{Global Rotation Image Warping.}
After estimating the global rotation $\bar{R}$ from the 360° optical flow, we warp the target image $I_{t+1}$ by the rotation $\bar{R}$ to align it to the source image $I_{t}$. %
We apply global rotation warping based on the coarse to fine principle.
First, we estimate the rotation $\bar{R}$ directly from ERP optical flow to roughly align the image $I_{t+1}$ and compensate for camera ego-motion.
We then use the cubemap optical flow to estimate the residual rotation $\hat{R}$ to fine-tune the global rotation alignment.

\vspace{-1em}
\paragraph{Global Rotation Optical Flow Warping.}
To recover the final 360° optical flow field $\mathcal{F}$ from the icosahedron optical flow $\tilde{\mathcal{F}}$, the end points of $\tilde{\mathcal{F}}$ need to align with the pixel position in the input image $I_{t+1}$.
When estimating the optical flow, only the target image $I_{t+1}$ is rotated, so we now need to rotate the end points of $\tilde{\mathcal{F}}$, i.e $\tilde{\mathcal{F}}^\text{EP}$, in the opposite direction.
To `rotate' an ERP pixel, we first convert it\new{s ERP image coordinates $\mathbf{x}$} to Cartesian 3D coordinates using \new{the inverse projection operator} $\mathcal{P}^{-1}$, rotate it about the centre of the sphere, and then convert it back to ERP \new{image coordinates} using $\mathcal{P}$.
Applied to optical flow, this yields:
\begin{equation}\label{equ:approach:globalwarp}
	\mathcal{F}(\mathbf{x})
	= \mathcal{F}^\text{EP}(\mathbf{x}) - \mathbf{x}
	= \mathcal{P} \left( \hat{R}^\top \cdot \bar{R}^\top \cdot \mathcal{P}^{-1}(\tilde{\mathcal{F}}^\text{EP}(\mathbf{x}))\right)  - \mathbf{x}
	\text{.}
\end{equation}

\section{Experiments}
\label{sec:exp}

We evaluate our method on the Replica \cite{StrauWMCWGEMRVCYBYPYZLCBGMPSBSNGLN2019} and OmniPhotos \cite{BerteYLR2020} datasets and achieve state-of-the-art performance on both datasets.
To evaluate the accuracy of 360° optical flow regardless projection, and to account for the wrap-around (\cref{sec:approach:definition}), we measure the \emph{spherical} average endpoint error (SEPE), \emph{spherical} average angle error (SAAE), and the \emph{spherical} root mean square error (SRMS).
Our single-threaded Python implementation computes 360° optical flow at 1280$\times$640 resolution in about 20 seconds.
We use DIS flow \cite{KroegTDV2016} for the tangent flow fields.
\looseness-1

\subsection{Datasets \& Error Metrics}
\label{sec:datasets-metrics}

We evaluate the performance of our method on synthetic and real-world datasets using multiple spherical error metrics.

\vspace{-1em}
\paragraph{Datasets.}
Unlike depth sensors, %
there is no feasible hardware for capturing high-quality optical flow from a real-world scene, especially for 360° optical flow.
For quantitative evaluation of optical flow, ground truth is necessary.
Inspired by \citet{ShugrLKLSSF2019}, we render ground-truth 360° images and optical flow from the reconstructed 3D mesh of a real-world scene.
The ground-truth data is rendered using the Replica dataset \cite{StrauWMCWGEMRVCYBYPYZLCBGMPSBSNGLN2019}, which not only contains high-quality HDR textures and 3D meshes, but also public code for a versatile renderer.
However, the official Replica rendering pipeline does not support any 360° camera model or optical flow generation.
We therefore implemented an ERP camera model using OpenGL geometry shaders that transform the 3D mesh from Cartesian coordinates to spherical coordinates, finally render the 3D mesh and ground-truth optical flow in the equirectangular format.
\looseness-1

We render three types of camera paths for ground-truth evaluation at a resolution of 1280$\times$640\,pixels.
A \emph{Circle}, in which the camera moves along a 50\,cm radius circle and faces outwards, with images rendered every 10°.
A \emph{Line}, where the camera faces the same direction and moves in a straight line, with images rendered every 20\,cm.
And \emph{Random}, with camera centres uniformly randomly sampled within a box of 1$\times$1$\times$1\,metres and rotations jittered by $\pm$10° degrees along each of the Euler angle axes.

Although the synthetic dataset is reasonably realistic, real-world data tends to be more challenging, with complex lighting and motions.
To further explore our method's performance qualitatively, we also test our method on the OmniPhotos dataset \cite{BerteYLR2020}, which contains diverse 360° videos of real scenes captured with a high-quality 360° video camera.

\vspace{-1em}
\paragraph{Error Metrics.}

Optical flow evaluation commonly uses the average angle error (AAE), end-point error (EPE), and root-mean-square error (RMS) metrics \cite{BakerSLRBS2011}.
To evaluate spherical 360° optical flow, we extend these metrics to the spherical domain by mapping ERP pixel coordinates to spherical coordinates on the unit sphere, and measuring geodesic distances instead of Euclidean distances in image space as used for perspective flow metrics.
This overcomes both distortions due to the equirectangular projection and also the wrap-around on the sphere.
Specifically, we propose the spherical AAE (SAAE), spherical EPE (SEPE, see \cref{equ:exp:epesph}) and spherical RMS (SRMS) metrics to evaluate 360° optical flow.
The spherical AAE (SAAE) measures the angle between the estimated and ground-truth optical flow vectors on the surface of a unit sphere using spherical trigonometry.
For SEPE and SRMS, $d(\cdot, \cdot)$ is the geodesic distance between endpoints on the unit sphere:
\begin{equation}\label{equ:exp:epesph}
	\text{SEPE} = \frac{1}{\new{|\Omega|}} \sum_{i \in \Omega} d\left( (\theta^\text{est}_i, \phi^\text{est}_i), (\theta^\text{GT}_i, \phi^\text{GT}_i)\right)
	\text{.}
\end{equation}

\subsection{Comparison}

We compare our method with OmniFlowNet \cite{ArtizZAD2020}, a state-of-the-art 360° optical flow method, as well as the following state-of-the-art perspective methods: RAFT~\cite{TeedD2020a}, PWC-Net~\cite{SunYLK2018} and DIS~\cite{KroegTDV2016}.
For \footurl{https://github.com/COATZ/OmniFlowNet}{OmniFlowNet}, \footurl{https://github.com/princeton-vl/RAFT}{RAFT} and \footurl{https://github.com/NVlabs/PWC-Net}{PWC-Net}, we use the official released code, and for DIS, we use OpenCV's implementation.
OmniFlowNet ran out of GPU memory on our NVIDIA RTX 2060 (6 GB RAM), so we downscale the input images to a resolution of 1024$\times$512\,pixels, and upsample the estimated flow fields back to the original image resolution afterwards.
The performance evaluation comprises quantitative and qualitative evaluation on the Replica 360° and OmniPhotos datasets.

\vspace{-1em}
\paragraph{Quantitative Evaluation.}

We show a quantitative comparison on the Replica 360° dataset using both perspective and spherical optical flow evaluation metrics in \cref{tab:exp:oferrorquality}.
The table
shows that our method achieves the best performance across the board.
All methods tend to be better on the \emph{Circle} paths than \emph{Line} paths, as the latter contains larger displacements that are more difficult to estimate accurately.
Meanwhile, our global rotation warping pre-aligns the input images, so that our method more easily handles camera ego-motion.

\begin{table}[h!]
	\centering
	\caption{\label{tab:exp:oferrorquality}%
		Optical flow errors on the synthetic Replica 360° dataset.
		See \cref{sec:datasets-metrics} for metrics.
	}
	\resizebox{0.85\linewidth}{!}{%
	\begin{tabular}{p{0.2cm}p{2.0cm}p{1.1cm}p{1.1cm}p{1.1cm}p{1.3cm}p{1.3cm}p{1.3cm}}
		\toprule
		&  Method & ${EPE}\downarrow$ & $AAE\downarrow$ & $RMS\downarrow$ & $SEPE\downarrow$ & ${SAAE}\downarrow$ & ${SRMS}\downarrow$ \\
		\midrule
		\multirow{5}{*}{\rotatebox[origin=c]{90}{\emph{Circle}}}
		& \new{OmniFlowNet} &\new{15.12} & \new{0.2618} & \new{60.79} &    \new{0.06025} &    \new{0.1316} &    \new{0.3963} \\
		& PWC-Net &    \new{15.73} &    \new{0.2701} &    \new{61.50} &    \new{0.05403} &    \new{0.1406} &    \new{0.3693} \\
		& RAFT    &    \new{15.75} &    \new{0.2769} &    \new{63.34} &    \new{0.04619} &    \new{0.1461} &    \new{0.3349} \\
		& DIS     &    \new{16.31} &    \new{0.2818} &    \new{62.29} &    \new{0.05430} &    \new{0.1503} &    \new{0.3578} \\
		& Ours    & \bf\new{\phantom{0}3.507} & \bf\new{0.1694} & \bf\new{34.21} & \bf\new{0.005370} & \bf\new{0.03480} & \bf\new{0.01021} \\
		\midrule
		\multirow{5}{*}{\rotatebox[origin=c]{90}{\emph{Line}}}
		& \new{OmniFlowNet} &\new{30.64} & \new{0.2515} & \new{92.36} &    \new{0.1229} &    \new{0.1276} &    \new{0.5390} \\
		& PWC-Net &    \new{32.23} &    \new{0.2594} &    \new{93.27} &    \new{0.1259} &     \new{0.1358} &    \new{0.5414} \\
		& RAFT    &    \new{32.38} &    \new{0.2935} &    \new{97.46} &    \new{0.08787} &    \new{0.1642} &    \new{0.4180} \\
		& DIS     &    \new{36.06} &    \new{0.3357} &    \new{99.07} &    \new{0.1127} &    \new{0.2093} &    \new{0.4668} \\
		& Ours    & \bf\new{\phantom{0}5.839} & \bf\new{0.1971} & \bf\new{40.74} & \bf\new{0.01063} & \bf\new{0.05951} & \bf\new{0.02098} \\
		\midrule
		\multirow{5}{*}{\rotatebox[origin=c]{90}{\emph{\new{Random}}}}
		& \new{OmniFlowNet} &\new{40.50} & \new{0.3229} & \new{106.2} &    \new{0.1321} &    \new{0.1940} &    \new{0.5034} \\
		& \new{PWC-Net} &    \new{41.98} &    \new{0.3294} &    \new{106.9} &    \new{0.1438} &    \new{0.2118} &    \new{0.5203} \\
		& \new{RAFT}    &    \new{38.31} &    \new{0.2978} &    \new{104.2} &    \new{0.1143} &    \new{0.1766} &    \new{0.4557} \\
		& \new{DIS}     &    \new{47.29} &    \new{0.4203} &    \new{113.2} &    \new{0.1728} &    \new{0.3000} &    \new{0.5545} \\
		& \new{Ours}    & \bf\new{14.10} & \bf\new{0.2192} & \bf\new{\phantom{0}59.78} & \bf\new{0.02717} & \bf\new{0.08849} & \bf\new{0.05753} \\
		\midrule
		\multirow{5}{*}{\rotatebox[origin=c]{90}{\emph{All}}} %
		& \new{OmniFlowNet} &\new{28.76} & \new{0.2788} & \new{86.48} &    \new{0.1051} &    \new{0.1511} &    \new{0.4797} \\
		& PWC-Net &    \new{29.98} &    \new{0.2863} &    \new{87.23} &    \new{0.1079} &    \new{0.1627} &    \new{0.4770} \\
		& RAFT    &    \new{28.81} &    \new{0.2893} &    \new{88.34} &    \new{0.08278} &    \new{0.1623} &    \new{0.4029} \\
		& DIS     &    \new{33.22} &    \new{0.3459} &    \new{91.54} &    \new{0.1133} &    \new{0.2199} &    \new{0.4597} \\
		& Ours    & \bf\new{\phantom{0}7.701} & \bf\new{0.1946} & \bf\new{44.62} & \bf\new{0.01411} & \bf\new{0.06027} & \bf\new{0.02905} \\
		\bottomrule
	\end{tabular}}%
\end{table}

\newcommand{\labeledfig}[4]{
	\begin{tikzpicture}
		\draw node[name=micrograph] {\includegraphics[width=#2\textwidth]{#1}}; %
		\draw (micrograph.north west)  node[anchor=north west,yshift=-2,#4]{\textbf{\tiny{#3}}}; %
	\end{tikzpicture}
}

\begin{figure*}[hbt!]
	\centering
	\begin{tabular}{@{}c*{6}{@{\hspace{-9pt}}c}@{}}
		\labeledfig{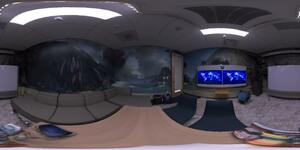}{0.165}{(a) $I_t$}{white} &
		\labeledfig{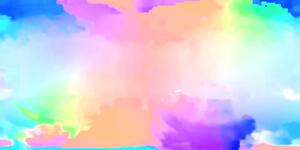}{0.165}{(b) DIS}{black} &
		\labeledfig{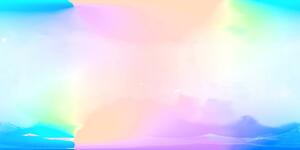}{0.165}{(c) Ours}{black} &
		\labeledfig{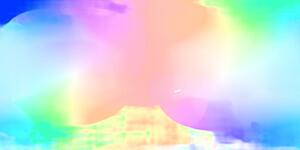}{0.165}{(d) PWC-Net}{black} &
		\labeledfig{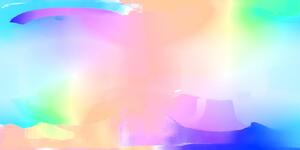}{0.165}{(e) RAFT}{black} &
		\labeledfig{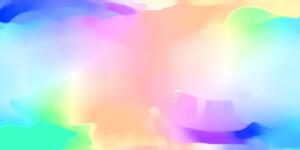}{0.165}{\new{(f) OmniFlowNet}}{black}
		\\[-1.0em]
		\labeledfig{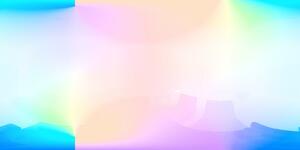}{0.165}{(g) GT}{black} &
		\labeledfig{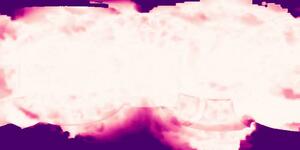}{0.165}{(h) DIS (SEPE)}{black} &
		\labeledfig{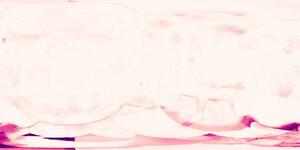}{0.165}{(i) Ours (SEPE)}{black} &
		\labeledfig{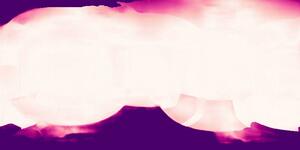}{0.165}{(j) PWC-Net (SEPE)}{black} &
		\labeledfig{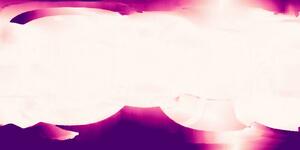}{0.165}{(k) RAFT (SEPE)}{black} &
		\labeledfig{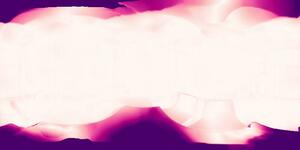}{0.165}{\new{(l) OmniFlowNet (SEPE)}}{black}
		\\[-0.9em]
		\labeledfig{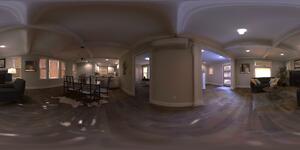}{0.165}{(a) $I_t$}{white} &
		\labeledfig{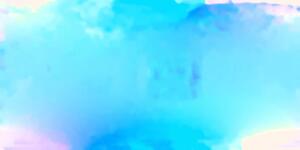}{0.165}{(b) DIS}{black} &
		\labeledfig{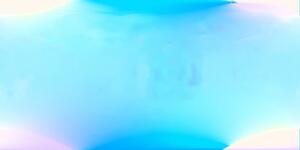}{0.165}{(c) Ours}{black} &
		\labeledfig{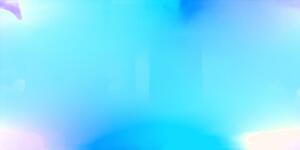}{0.165}{(d) PWC-Net}{black} &
		\labeledfig{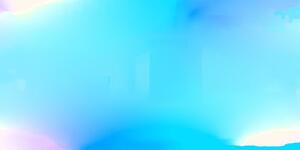}{0.165}{(e) RAFT}{black} &
		\labeledfig{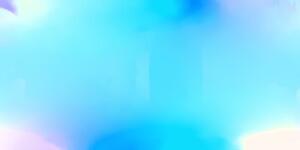}{0.165}{\new{(f) OmniFlowNet}}{black} 
		\\[-1.0em]
		\labeledfig{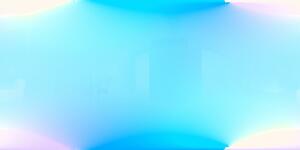}{0.165}{(g) GT}{black} &
		\labeledfig{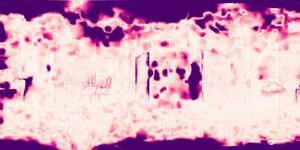}{0.165}{(h) DIS (SEPE)}{black} &
		\labeledfig{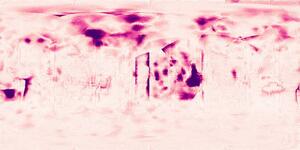}{0.165}{(i) Ours (SEPE)}{black} &
		\labeledfig{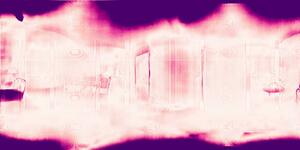}{0.165}{(j) PWC-Net (SEPE)}{black} &
		\labeledfig{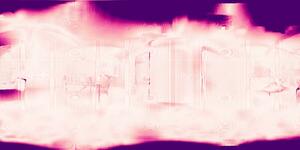}{0.165}{(k) RAFT (SEPE)}{black} &
		\labeledfig{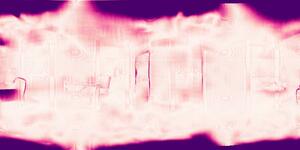}{0.165}{\new{(l) OmniFlowNet (SEPE)}}{black}
		\\[-0.9em]
		\labeledfig{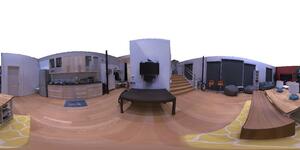}{0.165}{(a) $I_t$}{black} &
		\labeledfig{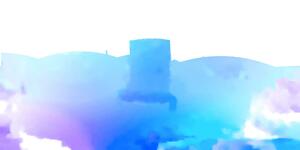}{0.165}{(b) DIS}{black} &
		\labeledfig{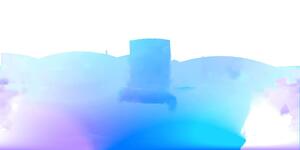}{0.165}{(c) Ours}{black} &
		\labeledfig{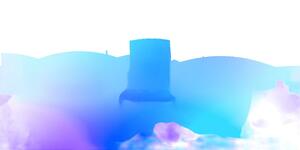}{0.165}{(d) PWC-Net}{black} &
		\labeledfig{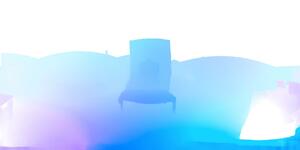}{0.165}{(e) RAFT}{black} &
		\labeledfig{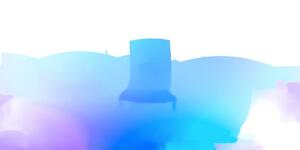}{0.165}{\new{(f) OmniFlowNet}}{black}
		\\[-1.0em]
		\labeledfig{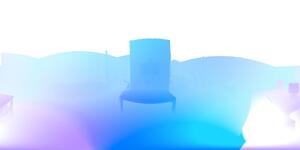}{0.165}{(g) GT}{black} &
		\labeledfig{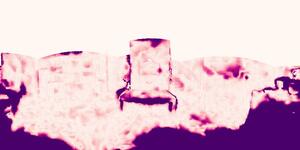}{0.165}{(h) DIS (SEPE)}{black} &
		\labeledfig{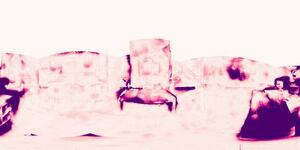}{0.165}{(i) Ours (SEPE)}{black} &
		\labeledfig{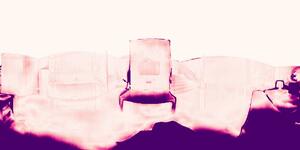}{0.165}{(j) PWC-Net (SEPE)}{black} &
		\labeledfig{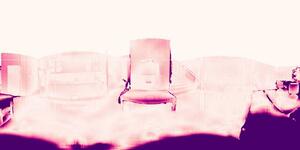}{0.165}{(k) RAFT (SEPE)}{black} &
		\labeledfig{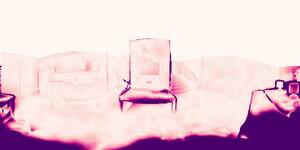}{0.165}{\new{(l) OmniFlowNet (SEPE)}}{black}
	\end{tabular}
	\caption{\label{fig:exp:compflow}%
		Estimated 360° optical flow and error heatmaps on the Replica 360° dataset:
		(a) Source image.
		(b–f) Estimated flow fields.
		(g) Ground-truth flow.
		(h–l) SEPE (spherical end-point error) heatmaps (lighter is better).
		Top: \emph{office\_0} scene, line camera motion.
		Middle: \emph{apartment\_0} scene, circle camera motion.
		\new{Bottom: \emph{frl\_apartment\_0} scene, random camera motion.}}
\end{figure*}

\noindent
\Cref{fig:exp:compflow} shows some example estimated 360° optical flow fields and their SEPE visualised as heatmaps (lighter is better).
Other methods mainly introduce errors on the top and bottom edges of the optical flow.
Our method has the lowest errors at the top and bottom, while the middle region is quite similar to the DIS result.

Based on the \cref{tab:exp:oferrorquality} and \cref{fig:exp:compflow}, we can conclude that
perspective optical flow cannot correctly handle the top and bottom regions of ERP images, but our method can help them improve dramatically in these areas.
Our method succeeds where the underlying optical flow method (DIS) fails, while maintaining the same performance in the equatorial image region.

\vspace{-1em}
\paragraph{Qualitative Evaluation.}

\begin{figure}[hbt!]
	\centering
	\begin{tabular}{@{}c*{6}{@{\hspace{-8pt}}c}@{}}
		\labeledfig{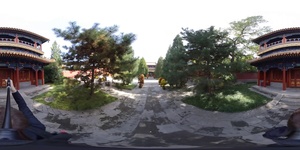}{0.165}{(a) $I_t$}{black} &
		\labeledfig{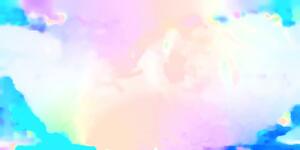}{0.165}{(b) DIS}{black} &
		\labeledfig{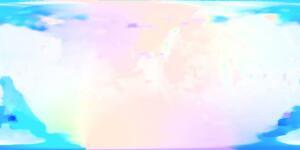}{0.165}{(c) Ours}{black} &
		\labeledfig{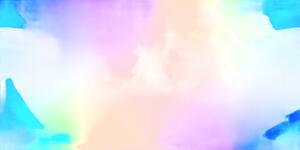}{0.165}{(d) PWC-Net}{black} &
		\labeledfig{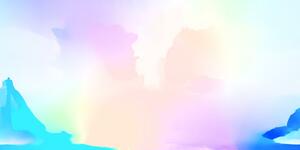}{0.165}{(e) RAFT}{black} &
		\labeledfig{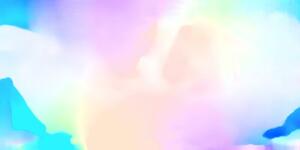}{0.165}{\new{(f) OmniFlowNet}}{black}
		\\[-1.0em]
		\labeledfig{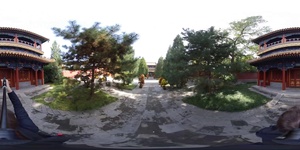}{0.165}{(g) $I_{t+1}$}{black} &
		\labeledfig{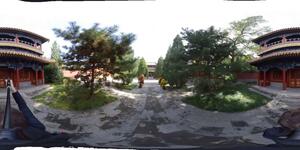}{0.165}{(h) DIS (Warp)}{black} &
		\labeledfig{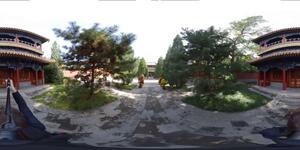}{0.165}{(i) Ours (Warp)}{black} &
		\labeledfig{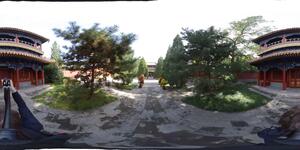}{0.165}{(j) PWC-Net (Warp)}{black} &
		\labeledfig{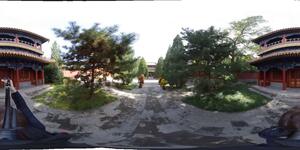}{0.165}{(k) RAFT (Warp)}{black} &
		\labeledfig{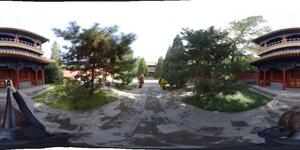}{0.165}{\new{(l) OmniFlowNet (Warp)}}{black}
		\\[-1.0em]
		&
		\labeledfig{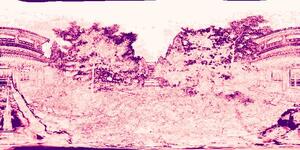}{0.165}{(m) DIS (Warp)}{black} &
		\labeledfig{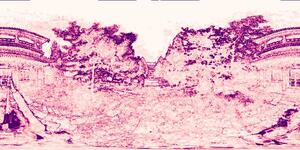}{0.165}{(n) Ours (Warp)}{black} &
		\labeledfig{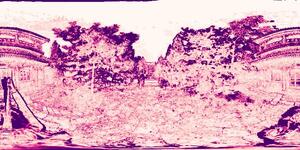}{0.165}{(o) PWC-Net (Warp)}{black} &
		\labeledfig{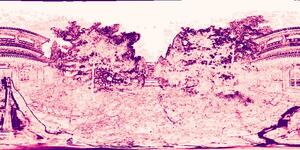}{0.165}{(p) RAFT (Warp)}{black} &
		\labeledfig{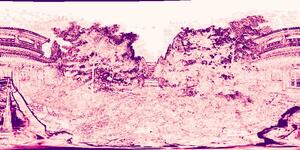}{0.165}{\new{(q) OmniFlowNet (Warp)}}{black} 
	\end{tabular}
	\caption{\label{fig:exp:backwardwarp}%
		Backward warping results on OmniPhotos' \emph{BeihaiPark} scene:
		(a, g) Source and target images.
		(b–f) Estimated flow fields.
		(h–l) Target image (g) warped to the source image (a) using the estimated optical flow (b–f).
		(m–q) Interpolation error heatmaps (lighter is better).\looseness-1}
\end{figure}

To analyse the performance of 360° optical flow methods on a real-world dataset without ground-truth optical flow, we measure the interpolation error \cite{BakerSLRBS2011}, i.e. the RGB colour difference between the source image and backward-warped target image.
\Cref{fig:exp:backwardwarp} shows an example result from the OmniPhotos dataset.
The error maps in \Cref{fig:exp:backwardwarp}(m–q) show less error at the top and bottom of images for our method's result.
The differences between methods is less pronounced than in the Replica 360° dataset, as the baseline between frames is smaller in OmniPhotos due to a higher video frame rate.

\begin{table}[h!]
	\centering
	\caption{\label{fig:exp:ablations}%
		The results of our ablation study. See \cref{sec:ablations} for details.}
	\resizebox{0.8\linewidth}{!}{%
		\begin{tabular}{p{0.2cm}p{1.9cm}p{1.1cm}p{1.1cm}p{1.1cm}p{1.3cm}p{1.3cm}p{1.3cm}}
			\toprule
			&  Method & ${EPE}\downarrow$ & $AAE\downarrow$ & $RMS\downarrow$ & $SEPE\downarrow$ & ${SAAE}\downarrow$ & ${SRMS}\downarrow$ \\
			\midrule
			\multirow{4}{*}{\rotatebox[origin=c]{90}{\emph{Circle}}}
			& Full method  &     \new{3.507} &     \new{0.1694} &     \new{34.21} & \bf \new{0.005370} &     \new{0.03480} &     \new{0.01021 } \\ 
			& w/o weight   &     \new{3.616} &     \new{0.1692} &     \new{34.86} &     \new{0.005405} &     \new{0.03478} &     \new{0.01235 } \\
			& w/o ERP      & \bf \new{3.434} & \bf \new{0.1688} & \bf \new{33.81} &     \new{0.005376} & \bf \new{0.03437} &     \new{0.009951} \\ 
			& w/o cubemap  &     \new{3.533} &     \new{0.1698} &     \new{34.04} &     \new{0.005399} &     \new{0.03519} & \bf \new{0.009730} \\ 
			& w/o ico      &     \new{3.481} &     \new{0.1711} &     \new{30.45} &     \new{0.006040} &     \new{0.03724} &     \new{0.01412 } \\
			\midrule
			\multirow{4}{*}{\rotatebox[origin=c]{90}{\emph{Line}}} 
			& Full method  &     \new{5.839} &     \new{0.1971} &     \new{40.74} &     \new{0.01063} &     \new{0.05951} &     \new{0.02098}  \\
			& w/o weight   &     \new{5.475} &     \new{0.1960} &     \new{32.36} &     \new{0.01100} & \bf \new{0.05815} &     \new{0.02214}   \\ 
			& w/o ERP      &     \new{5.749} & \bf \new{0.1945} &     \new{40.11} &     \new{0.01069} &     \new{0.05881} &     \new{0.02107}  \\ 
			& w/o cubemap  &     \new{5.816} &     \new{0.1953} &     \new{40.09} &     \new{0.01075} &     \new{0.05825} &     \new{0.02136}  \\ 
			& w/o ico      & \bf \new{4.262} &     \new{0.2042} & \bf \new{26.28} & \bf \new{0.01015} &     \new{0.07242} & \bf \new{0.01818}  \\ 
			\midrule
			\multirow{4}{*}{\rotatebox[origin=c]{90}{\emph{Random}}} 
			& Full method  &     \new{14.10} &     \new{0.2192} &     \new{59.78} &     \new{0.02717} &     \new{0.08849} &     \new{0.05753} \\
			& w/o weight   & \bf \new{13.66} &     \new{0.2176} &     \new{57.87} &     \new{0.02723} &     \new{0.08679} &     \new{0.05684}  \\ 
			& w/o ERP      &     \new{15.71} &     \new{0.2211} &     \new{60.81} &     \new{0.03212} &     \new{0.09007} &     \new{0.06161} \\ 
			& w/o cubemap  &     \new{13.96} & \bf \new{0.2145} &     \new{58.92} & \bf \new{0.02652} & \bf \new{0.08428} & \bf \new{0.05619} \\ 
			& w/o ico      &     \new{14.81} &     \new{0.2490} & \bf \new{57.15} &     \new{0.03389} &     \new{0.1167 } &     \new{0.06173} \\ 
			\midrule
			\multirow{4}{*}{\rotatebox[origin=c]{90}{\emph{All}}}
			& Full method  &     \new{7.701} &     \new{0.1946} &     \new{44.62} & \bf \new{0.01411} &     \new{0.06027} & \bf \new{0.02905}   \\
			& w/o weight   &     \new{7.585} &     \new{0.1942} &     \new{41.70} &     \new{0.01455} &     \new{0.05991} &     \new{0.03045}  \\
			& w/o ERP      &     \new{8.298} &     \new{0.1948} &     \new{44.91} &     \new{0.01606} &     \new{0.06109} &     \new{0.03088} \\
			& w/o cubemap  &     \new{7.771} & \bf \new{0.1932} &     \new{44.35} &     \new{0.01422} & \bf \new{0.05924} &     \new{0.02909} \\
			& w/o ico      & \bf \new{7.519} &     \new{0.2081} & \bf \new{37.96} &     \new{0.01670} &     \new{0.07547} &     \new{0.03134} \\ 
			\bottomrule
	\end{tabular}}%
\end{table}%

\subsection{Ablation Studies}
\label{sec:ablations}

\paragraph{Global Rotation Warping.}
The multi-step rotation warping aligns image $I_{t+1}$ to $I_t$ by estimating rotation using the ERP and cubemap optical flow.
The results without ERP optical flow alignment (`w/o ERP') and without cubemap optical flow alignment (`w/o cubemap') are shown in \cref{fig:exp:ablations}.
\new{In different type camera motion data different rotation gets best performance.
But `Full method'make most of and balance each term's advantage, to make it get the best performance on `\emph{All}' data.}

\vspace{-1em}
\paragraph{Stitch Blending Weight.}
The blending weight aims to smoothly and consistently stitch the overlap area of different faces' optical flow.
The result without blending weight is shown as `w/o weight' in \cref{fig:exp:ablations}, which replaces all weights with the unit weight.
\new{The `Full Method' outperforms the `w/o weight' because the blending weight brings down the unreliable optical flow, such as the pixels move out or in the tangent image boundary but their reliable optical flow can get from corresponding pixels on other tangent images.}

\section{Conclusion}

We proposed a flexible method for estimating 360° optical flow using tangent images and global rotation warping.
Tangent images use gnomonic projection to uniformly sample a 360° ERP image from an initial cubemap to the final icosahedron-based tangent images to overcome the equirectangular projection distortions.
Any existing off-the-shelf optical flow method can be used on the tangent images.
Our method uses global rotation warping to pre-align the input images and overcome large motions between frames.
Finally, we blend optical flow to reduce discontinuities at face boundaries.
Our evaluation shows that our method achieves state-of-the-art performance for 360° optical flow estimation.

\new{Our proposed approach overcomes the limited resolutions supported by CNN-based methods due to model size constraints (e.g. 1024$\times$512 \cite{ArtizZAD2020}) by processing tangent images separately.
This enables the estimation of high-resolution 360° optical flow, for example at 2K$\times$1K resolution, which is not supported by existing methods.}

\vspace{-1em}
\new{\paragraph{Limitations.}
Large pose changes are a common limitation of most optical flow methods, which are primarily designed for handling small motions.
In our method, we use global rotation alignment to effectively minimise the net disparity across the unit sphere.
This helps in the case of moderate pose changes, but large pose changes remain challenging.
}

\vspace{-1em}
\paragraph{Future Work.}
Our method struggles with large camera translations; one could replace the global rotation warping with a meshgrid-based warping method that can pre-align both rotational and translational camera motions.

\section*{Acknowledgements}
We thank the reviewers for their valuable feedback that has helped us improve our paper.
This work was supported by an EPSRC-UKRI Innovation Fellowship (EP/S001050/1) and RCUK grant CAMERA (EP/M023281/1, EP/T022523/1).

\clearpage

\begin{appendices}
\section{Padding Size}
\label{sec:sup:ablations}

\begin{figure*}[hbt!]
	\centering
	\includegraphics[width=0.80\textwidth]{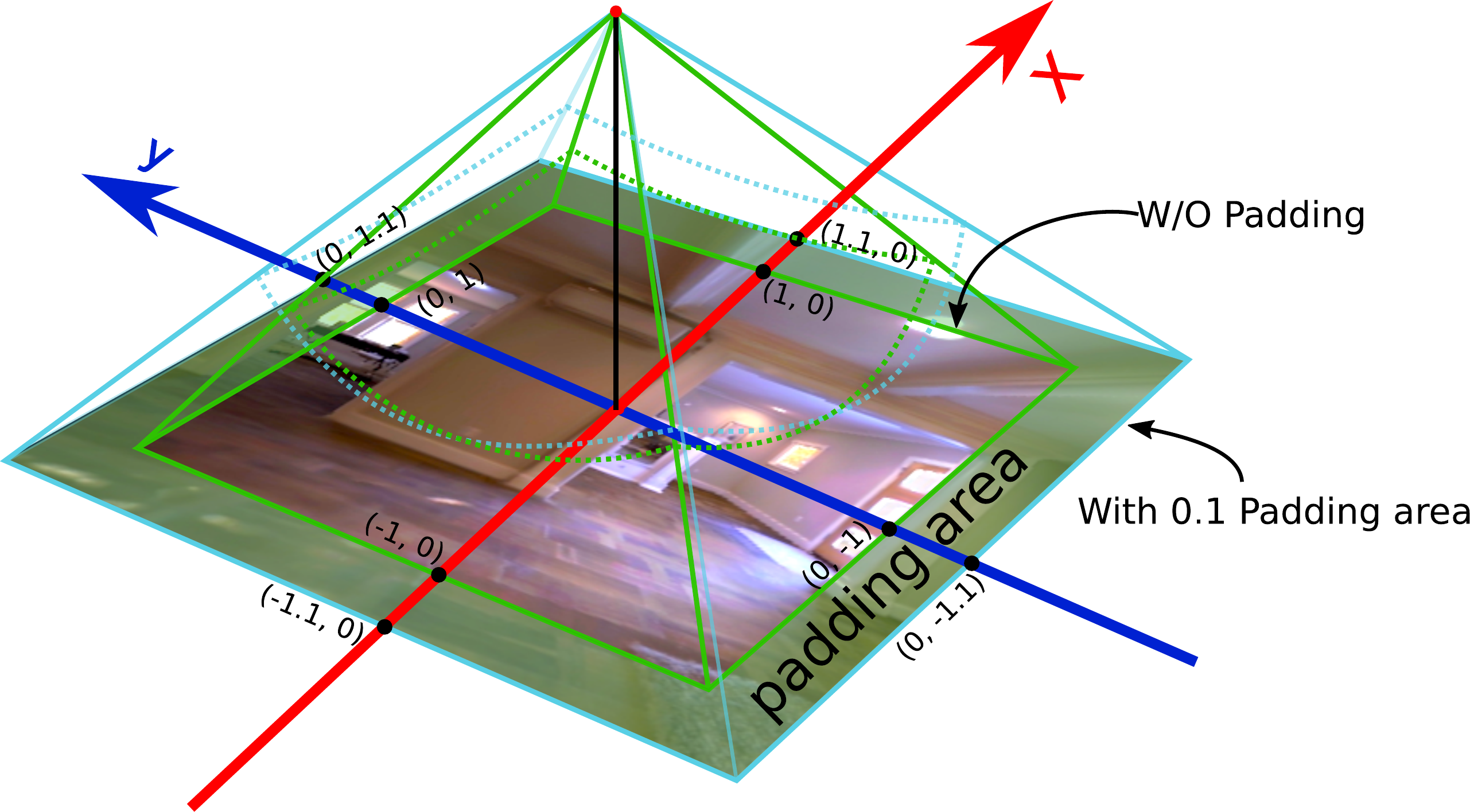}
	\caption{\label{fig:sup:howtopadding}%
		The tangent image padding.}
\end{figure*}

\begin{figure*}[hbt!]
	\centering
	\includegraphics[width=0.80\textwidth]{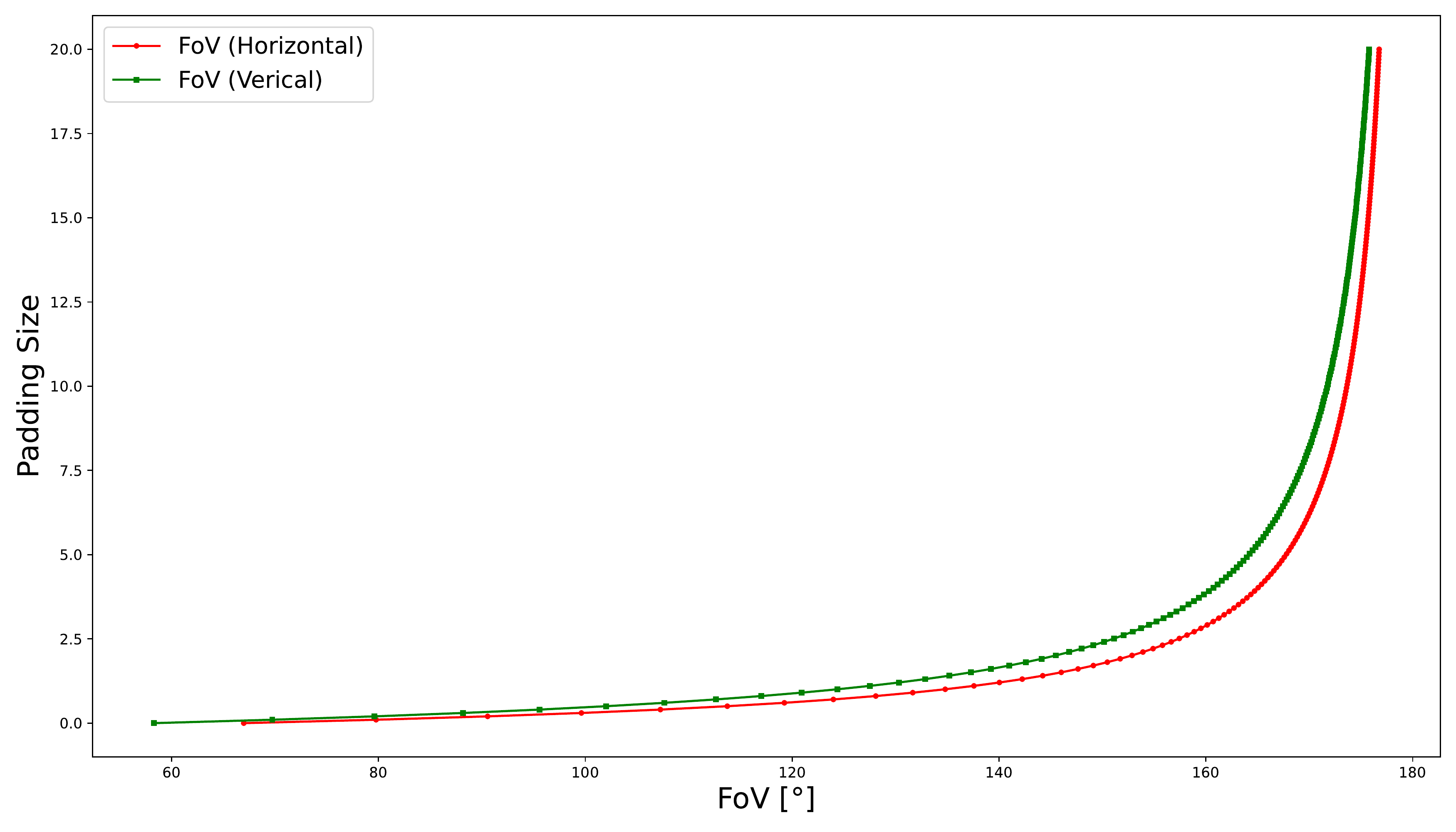}
	\caption{\label{fig:sup:paddingvsfov}%
		The relationship between tangent image FoV and padding size. The red and green curves are the tangent image horizontal and vertical FoV, respectively.}
\end{figure*}

\begin{figure*}[hbt!]
	\centering
	\includegraphics[width=0.80\textwidth]{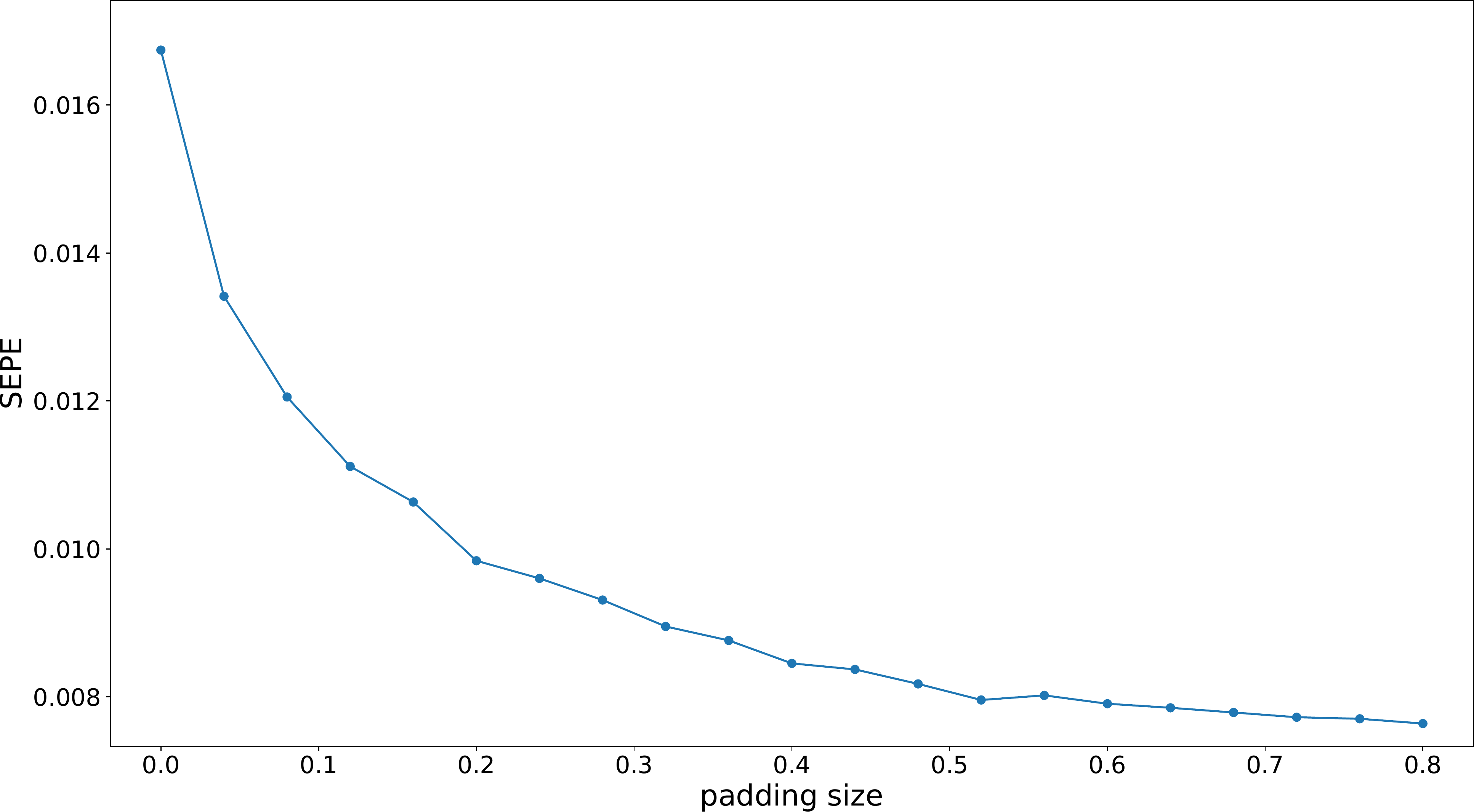}
	\caption{\label{fig:sup:ablationpadding}%
		The spherical end-point error (SEPE) for a range of padding sizes.}
\end{figure*}

The padding size is how much expand on the gnomonic projection coordinate system.
When the padding size is 0.1, shown in \cref{fig:sup:howtopadding}, the tangent area is expended from 1.0 to 1.1 ($1.0  + padding\_size$).
The tangent image padding extends the field of view  (see Section 3.2 in the main paper), such that there is more overlap between the source and target images to find better correspondences and improve optical flow consistency.
However, increased padding size reduces the tangent image angular resolution. 
As shown in the \cref{fig:sup:paddingvsfov}, the image's field of view (FoV) increasing with raising padding size.
We test padding size from 0.0 to 0.6 and show the result in \cref{fig:sup:ablationpadding}.
Larger padding sizes reduce the spherical end-point error (SEPE).
The SEPE plateaus at a padding of around 0.5, which optimally trades off angular resolution and padding size.

\section{Additional Results}

Here we show additional results for quantitative and qualitative evaluation.

\paragraph{Quantitative Evaluation.}

\begin{figure*}[hbt!]
	\centering
	\begin{tabular}{@{}c*{6}{@{\hspace{-9pt}}c}@{}}
		\multicolumn{5}{l}{\footnotesize{\emph{Circle}:\emph{apartment\_0}}}
		\\[-0.6em]
		\labeledfig{images/of_error/apartment_0_circ/apartment_0_circ_1k_0_0004_rgb_pano.jpg}{0.165}{(a) $I_t$}{white} &
		\labeledfig{images/of_error/apartment_0_circ/apartment_0_circ_1k_0_dis_0004_opticalflow_backward_pano.flo_flow_vis.jpg}{0.165}{(b) $DIS$}{black} &
		\labeledfig{images/of_error/apartment_0_circ/apartment_0_circ_1k_0_our_0.4_0004_opticalflow_backward_pano.flo_flow_vis.jpg}{0.165}{(c) $Our$}{black} &
		\labeledfig{images/of_error/apartment_0_circ/apartment_0_circ_1k_0_pwcnet_0004_opticalflow_backward_pano.flo_flow_vis.jpg}{0.165}{(d) $PWCNet$}{black} &
		\labeledfig{images/of_error/apartment_0_circ/apartment_0_circ_1k_0_raft_0004_opticalflow_backward_pano.flo_flow_vis.jpg}{0.165}{(e) $RAFT$}{black} &
		\labeledfig{images/of_error/apartment_0_circ/apartment_0_circ_1k_0_omniflownet_0004_opticalflow_backward_pano.flo_flow_vis.jpg}{0.165}{(f) $OmniFlowNet$}{black}
		\\[-1.0em]
		\labeledfig{images/of_error/apartment_0_circ/apartment_0_circ_1k_0_gt_0004_opticalflow_backward_pano.flo_flow_vis.jpg}{0.165}{(g) GT}{black} &
		\labeledfig{images/of_error/apartment_0_circ/apartment_0_circ_1k_0_dis_0004_opticalflow_backward_pano.flo_error_vis.jpg}{0.165}{(h) $DIS\_SEPE$}{white} &
		\labeledfig{images/of_error/apartment_0_circ/apartment_0_circ_1k_0_our_0.4_0004_opticalflow_backward_pano.flo_error_vis.jpg}{0.165}{(i) $Our\_SEPE$}{white} &
		\labeledfig{images/of_error/apartment_0_circ/apartment_0_circ_1k_0_pwcnet_0004_opticalflow_backward_pano.flo_error_vis.jpg}{0.165}{(j) $PWCNet\_SEPE$}{white} &
		\labeledfig{images/of_error/apartment_0_circ/apartment_0_circ_1k_0_raft_0004_opticalflow_backward_pano.flo_error_vis.jpg}{0.165}{(k) $RAFT\_SEPE$}{white} &
		\labeledfig{images/of_error/apartment_0_circ/apartment_0_circ_1k_0_omniflownet_0004_opticalflow_backward_pano.flo_error_vis.jpg}{0.165}{(l) $OmniFlowNet\_SEPE$}{white}
		\\[-0.6em]
		\multicolumn{5}{l}{\footnotesize{\emph{Circle}:\emph{apartment\_1}}}
		\\[-0.6em]
		\labeledfig{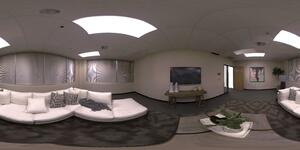}{0.165}{(a) $I_t$}{white} &
		\labeledfig{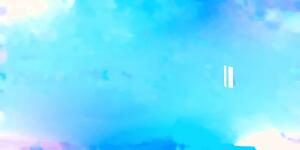}{0.165}{(b) $DIS$}{black} &
		\labeledfig{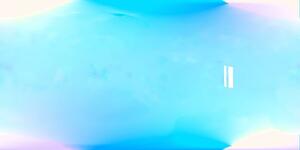}{0.165}{(c) $Our$}{black} &
		\labeledfig{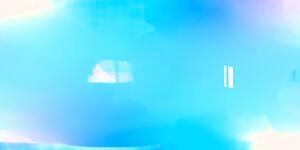}{0.165}{(d) $PWCNet$}{black} &
		\labeledfig{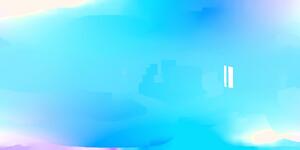}{0.165}{(e) $RAFT$}{black} &
		\labeledfig{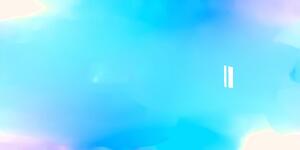}{0.165}{(f) $OmniFlowNet$}{black}
		\\[-1.0em]
		\labeledfig{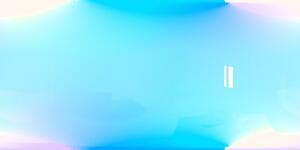}{0.165}{(g) GT}{black} &
		\labeledfig{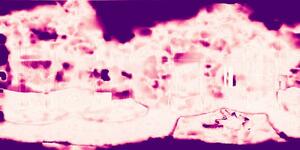}{0.165}{(h) $DIS\_SEPE$}{black} &
		\labeledfig{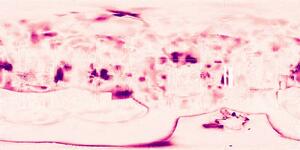}{0.165}{(i) $Our\_SEPE$}{black} &
		\labeledfig{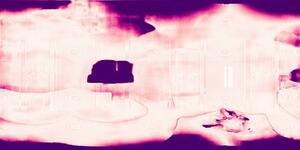}{0.165}{(j) $PWCNet\_SEPE$}{black} &
		\labeledfig{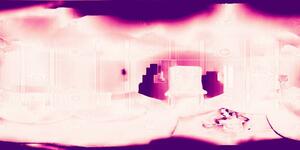}{0.165}{(k) $RAFT\_SEPE$}{black} &
		\labeledfig{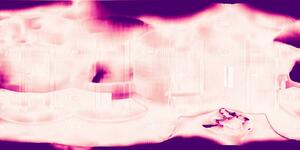}{0.165}{(l) $OmniFlowNet\_SEPE$}{black}
		\\[-0.6em]
		\multicolumn{5}{l}{\footnotesize{\emph{Circle}:\emph{frl\_apartment\_0}}}
		\\[-0.6em]
		\labeledfig{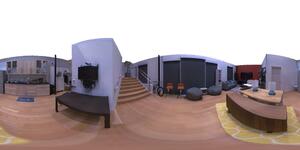}{0.165}{(a) $I_t$}{black} &
		\labeledfig{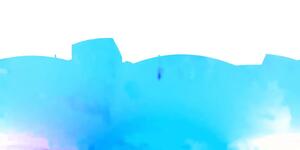}{0.165}{(b) $DIS$}{black} &
		\labeledfig{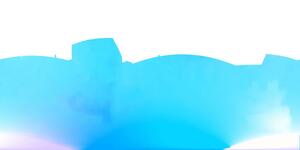}{0.165}{(c) $Our$}{black} &
		\labeledfig{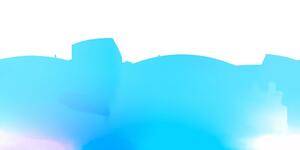}{0.165}{(d) $PWCNet$}{black} &
		\labeledfig{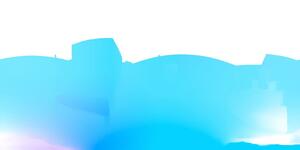}{0.165}{(e) $RAFT$}{black} &
		\labeledfig{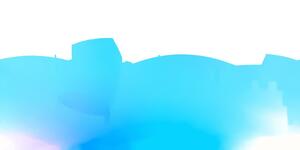}{0.165}{(f) $OmniFlowNet$}{black}
		\\[-1.0em]
		\labeledfig{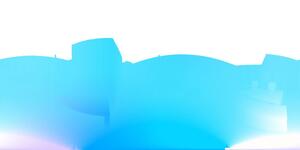}{0.165}{(g) GT}{black} &
		\labeledfig{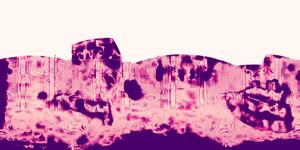}{0.165}{(h) $DIS\_SEPE$}{black} &
		\labeledfig{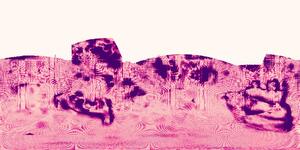}{0.165}{(i) $Our\_SEPE$}{black} &
		\labeledfig{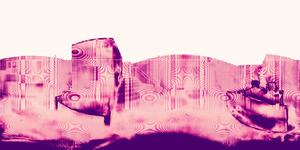}{0.165}{(j) $PWCNet\_SEPE$}{black} &
		\labeledfig{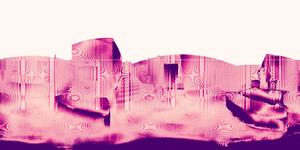}{0.165}{(k) $RAFT\_SEPE$}{black} &
		\labeledfig{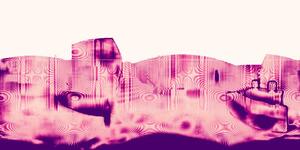}{0.165}{(l) $OmniFlowNet\_SEPE$}{black}
		\\[-0.6em]
		\multicolumn{5}{l}{\footnotesize{\emph{Circle}:\emph{hotel\_0}}}
		\\[-0.6em]
		\labeledfig{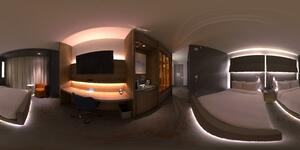}{0.165}{(a) $I_t$}{white} &
		\labeledfig{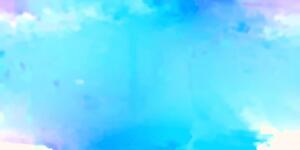}{0.165}{(b) $DIS$}{black} &
		\labeledfig{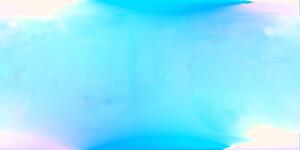}{0.165}{(c) $Our$}{black} &
		\labeledfig{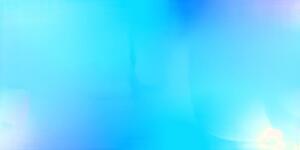}{0.165}{(d) $PWCNet$}{black} &
		\labeledfig{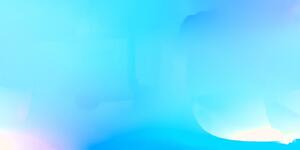}{0.165}{(e) $RAFT$}{black} &
		\labeledfig{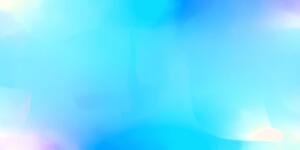}{0.165}{(f) $OmniFlowNet$}{black} &
		\\[-1.0em]
		\labeledfig{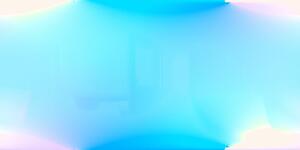}{0.165}{(g) GT}{black} &
		\labeledfig{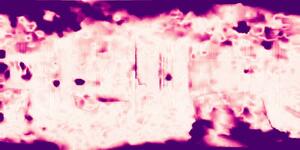}{0.165}{(h) $DIS\_SEPE$}{black} &
		\labeledfig{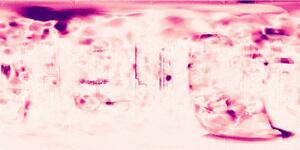}{0.165}{(i) $Our\_SEPE$}{black} &
		\labeledfig{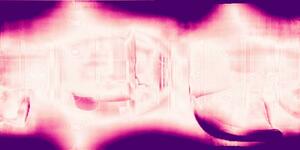}{0.165}{(j) $PWCNet\_SEPE$}{black} &
		\labeledfig{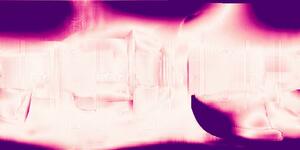}{0.165}{(k) $RAFT\_SEPE$}{black} &
		\labeledfig{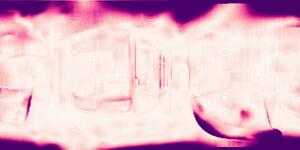}{0.165}{(l) $OmniFlowNet\_SEPE$}{black} 
		\\[-0.6em]
		\multicolumn{5}{l}{\footnotesize{\emph{Circle}:\emph{office\_1}}}
		\\[-0.6em]
		\labeledfig{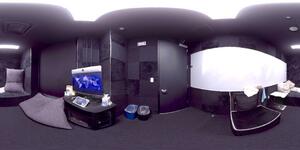}{0.165}{(a) $I_t$}{white} &
		\labeledfig{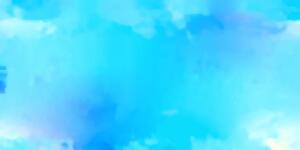}{0.165}{(b) $DIS$}{black} &
		\labeledfig{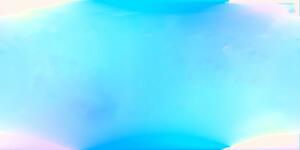}{0.165}{(c) $Our$}{black} &
		\labeledfig{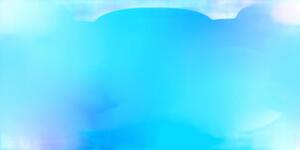}{0.165}{(d) $PWCNet$}{black} &
		\labeledfig{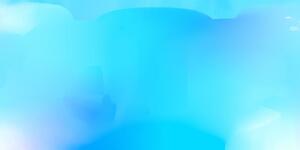}{0.165}{(e) $RAFT$}{black} &
		\labeledfig{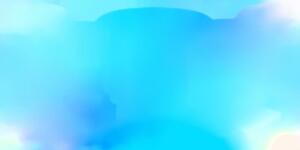}{0.165}{(f) $OmniFlowNet$}{black}
		\\[-1.0em]
		\labeledfig{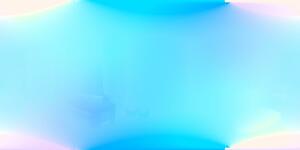}{0.165}{(g) GT}{black} &
		\labeledfig{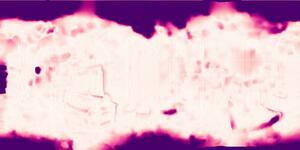}{0.165}{(h) $DIS\_SEPE$}{black} &
		\labeledfig{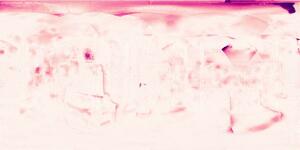}{0.165}{(i) $Our\_SEPE$}{black} &
		\labeledfig{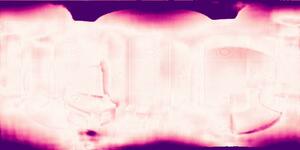}{0.165}{(j) $PWCNet\_SEPE$}{black} &
		\labeledfig{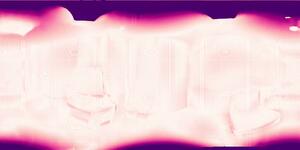}{0.165}{(k) $RAFT\_SEPE$}{black} &
		\labeledfig{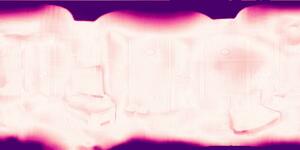}{0.165}{(l) $OmniFlowNet\_SEPE$}{black}
		\\[-0.6em]
		\multicolumn{5}{l}{\footnotesize{\emph{Circle}:\emph{room\_0}}}
		\\[-0.6em]
		\labeledfig{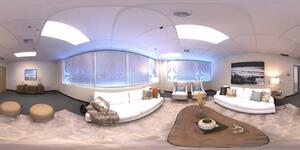}{0.165}{(a) $I_t$}{black} &
		\labeledfig{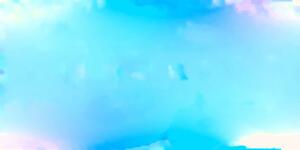}{0.165}{(b) $DIS$}{black} &
		\labeledfig{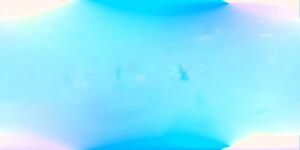}{0.165}{(c) $Our$}{black} &
		\labeledfig{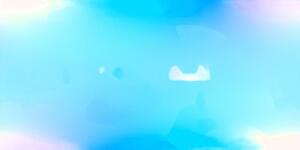}{0.165}{(d) $PWCNet$}{black} &
		\labeledfig{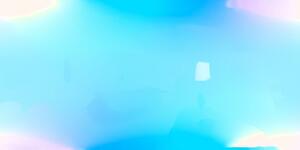}{0.165}{(e) $RAFT$}{black} &
		\labeledfig{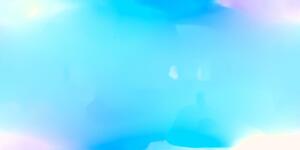}{0.165}{(f) $OmniFlowNet$}{black}
		\\[-1.0em]
		\labeledfig{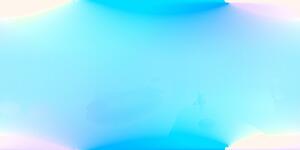}{0.165}{(g) GT}{black} &
		\labeledfig{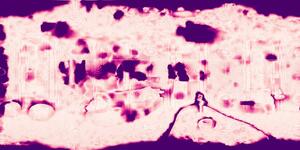}{0.165}{(h) $DIS\_SEPE$}{black} &
		\labeledfig{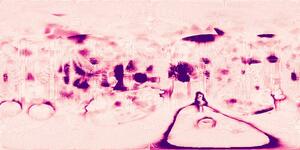}{0.165}{(i) $Our\_SEPE$}{black} &
		\labeledfig{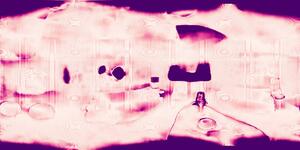}{0.165}{(j) $PWCNet\_SEPE$}{black} &
		\labeledfig{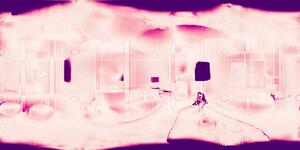}{0.165}{(k) $RAFT\_SEPE$}{black} &
		\labeledfig{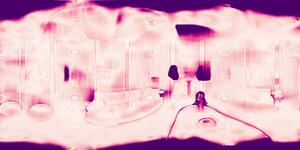}{0.165}{(l) $OmniFlowNet\_SEPE$}{black}
		\\[-0.6em]
		\multicolumn{5}{l}{\footnotesize{\emph{Circle}:\emph{room\_1}}}
		\\[-0.6em]
		\labeledfig{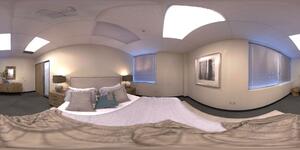}{0.165}{(a) $I_t$}{black} &
		\labeledfig{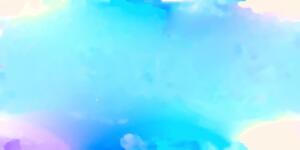}{0.165}{(b) $DIS$}{black} &
		\labeledfig{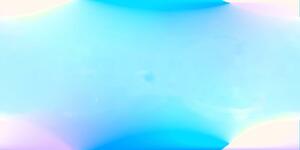}{0.165}{(c) $Our$}{black} &
		\labeledfig{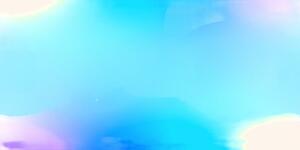}{0.165}{(d) $PWCNet$}{black} &
		\labeledfig{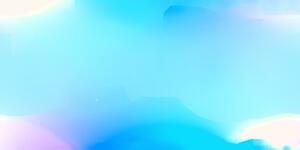}{0.165}{(e) $RAFT$}{black} &
		\labeledfig{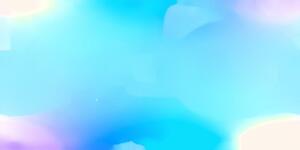}{0.165}{(f) $OmniFlowNet$}{black}	
		\\[-1.0em]
		\labeledfig{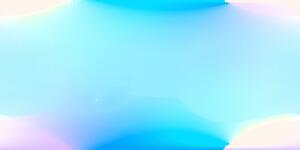}{0.165}{(g) GT}{black} &
		\labeledfig{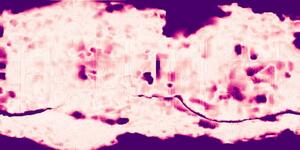}{0.165}{(h) $DIS\_SEPE$}{black} &
		\labeledfig{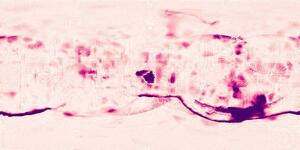}{0.165}{(i) $Our\_SEPE$}{black} &
		\labeledfig{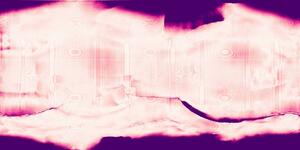}{0.165}{(j) $PWCNet\_SEPE$}{black} &
		\labeledfig{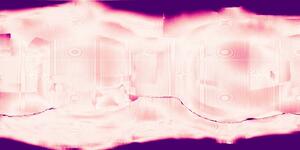}{0.165}{(k) $RAFT\_SEPE$}{black} &
		\labeledfig{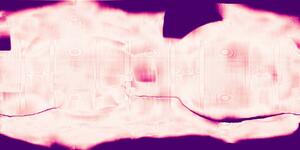}{0.165}{(l) $OmniFlowNet\_SEPE$}{black}
\end{tabular}
\caption{\label{fig:sup:compflowcircle}%
	Estimated 360° optical flow and error heatmaps on the Replica 360° dataset (\emph{Circle}).}
\end{figure*}

\begin{figure*}[hbt!]
	\centering
	\begin{tabular}{@{}c*{6}{@{\hspace{-9pt}}c}@{}}
		\\[-0.5em]
		\multicolumn{5}{l}{\footnotesize{\emph{Line}:\emph{apartment\_0}}}
		\\[-0.5em]
		\labeledfig{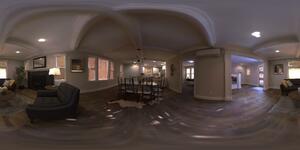}{0.165}{(a) $I_t$}{white} &
		\labeledfig{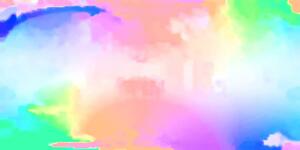}{0.165}{(b) $DIS$}{black} &
		\labeledfig{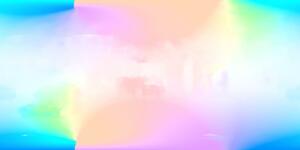}{0.165}{(c) $Our$}{black} &
		\labeledfig{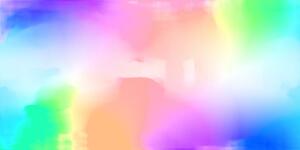}{0.165}{(d) $PWCNet$}{black} &
		\labeledfig{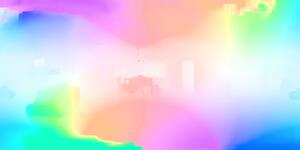}{0.165}{(e) $RAFT$}{black} &
		\labeledfig{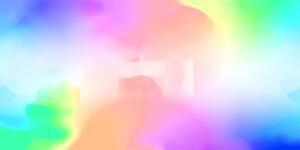}{0.165}{(f) $OmniFlowNet$}{black} 
		\\[-1.0em]
		\labeledfig{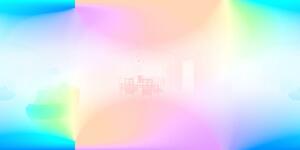}{0.165}{(g) GT}{black} &
		\labeledfig{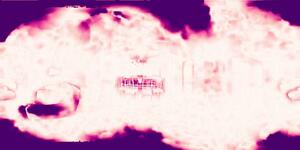}{0.165}{(h) $DIS\_SEPE$}{black} &
		\labeledfig{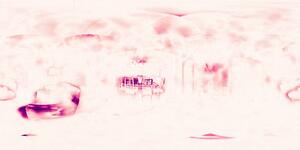}{0.165}{(i) $Our\_SEPE$}{black} &
		\labeledfig{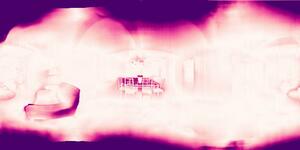}{0.165}{(j) $PWCNet\_SEPE$}{black} &
		\labeledfig{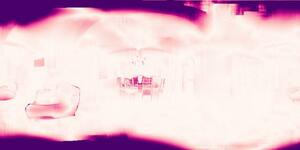}{0.165}{(k) $RAFT\_SEPE$}{black} &
		\labeledfig{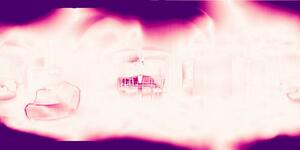}{0.165}{(l) $OmniFlowNet\_SEPE$}{black}
		\\[-0.6em]
		\multicolumn{5}{l}{\footnotesize{\emph{Line}:\emph{apartment\_1}}}
		\\[-0.5em]
		\labeledfig{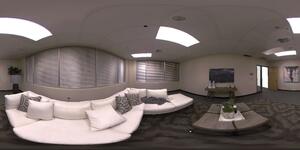}{0.165}{(a) $I_t$}{black} &
		\labeledfig{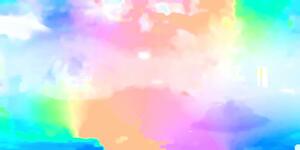}{0.165}{(b) $DIS$}{black} &
		\labeledfig{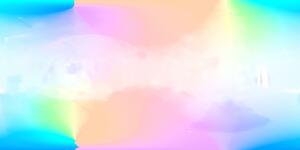}{0.165}{(c) $Our$}{black} &
		\labeledfig{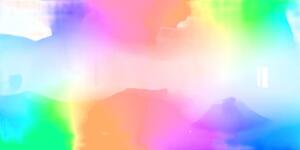}{0.165}{(d) $PWCNet$}{black} &
		\labeledfig{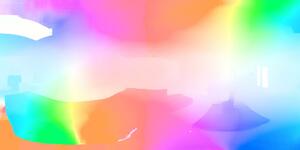}{0.165}{(e) $RAFT$}{black} &
		\labeledfig{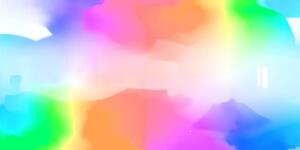}{0.165}{(f) $OmniFlowNet$}{black}
		\\[-1.0em]
		\labeledfig{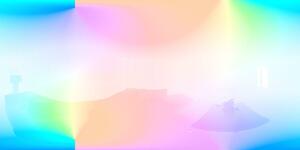}{0.165}{(g) GT}{black} &
		\labeledfig{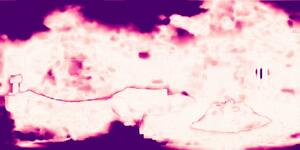}{0.165}{(h) $DIS\_SEPE$}{black} &
		\labeledfig{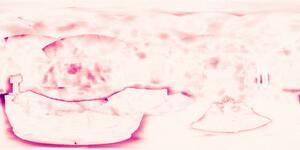}{0.165}{(i) $Our\_SEPE$}{black} &
		\labeledfig{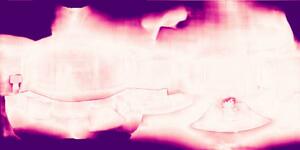}{0.165}{(j) $PWCNet\_SEPE$}{black} &
		\labeledfig{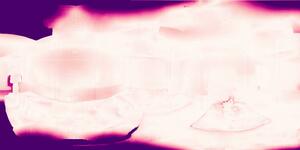}{0.165}{(k) $RAFT\_SEPE$}{black} &
		\labeledfig{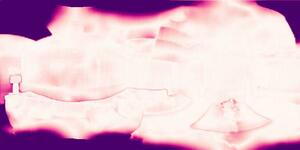}{0.165}{(l) $OmniFlowNet\_SEPE$}{black}
		\\[-0.6em]
		\multicolumn{5}{l}{\footnotesize{\emph{Line}:\emph{frl\_apartment\_1}}}
		\\[-0.6em]
		\labeledfig{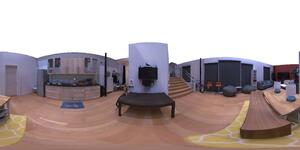}{0.165}{(a) $I_t$}{white} &
		\labeledfig{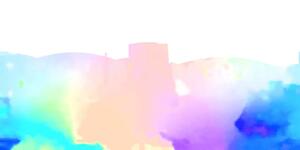}{0.165}{(b) $DIS$}{black} &
		\labeledfig{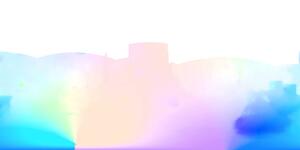}{0.165}{(c) $Our$}{black} &
		\labeledfig{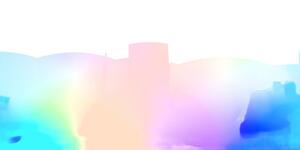}{0.165}{(d) $PWCNet$}{black} &
		\labeledfig{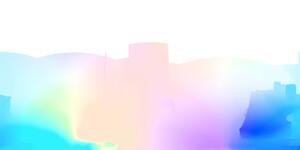}{0.165}{(e) $RAFT$}{black} &
		\labeledfig{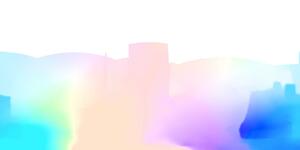}{0.165}{(f) $OmniFlowNet$}{black}
		\\[-1.0em]
		\labeledfig{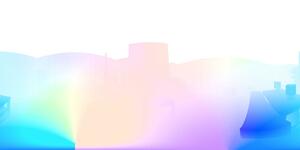}{0.165}{(g) GT}{black} &
		\labeledfig{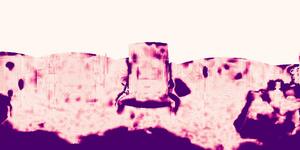}{0.165}{(h) $DIS\_SEPE$}{black} &
		\labeledfig{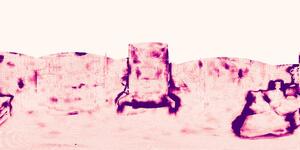}{0.165}{(i) $Our\_SEPE$}{black} &
		\labeledfig{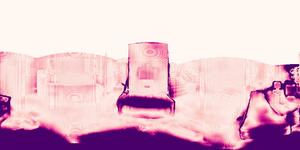}{0.165}{(j) $PWCNet\_SEPE$}{black} &
		\labeledfig{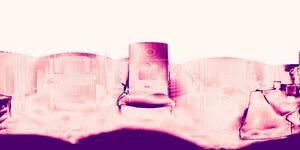}{0.165}{(k) $RAFT\_SEPE$}{black} &
		\labeledfig{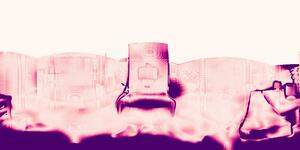}{0.165}{(l) $OmniFlowNet$}{black}
		\\[-0.6em]
		\multicolumn{5}{l}{\footnotesize{\emph{Line}:\emph{hotel\_0}}}
		\\[-0.6em]
		\labeledfig{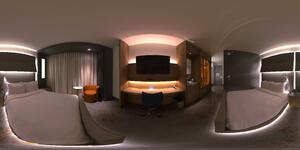}{0.165}{(a) $I_t$}{white} &
		\labeledfig{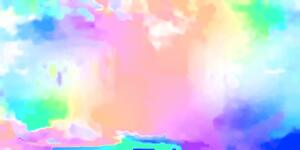}{0.165}{(b) $DIS$}{black} &
		\labeledfig{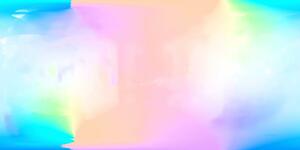}{0.165}{(c) $Our$}{black} &
		\labeledfig{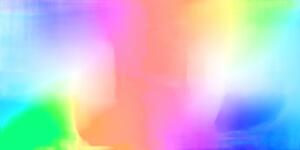}{0.165}{(d) $PWCNet$}{black} &
		\labeledfig{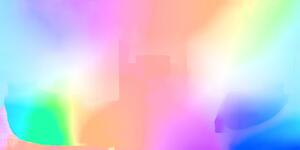}{0.165}{(e) $RAFT$}{black} &
		\labeledfig{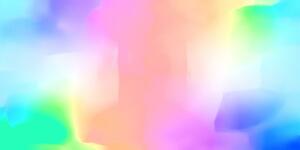}{0.165}{(f) $OmniFlowNet$}{black}
		\\[-1.0em]
		\labeledfig{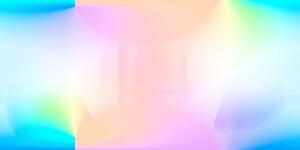}{0.165}{(g) GT}{black} &
		\labeledfig{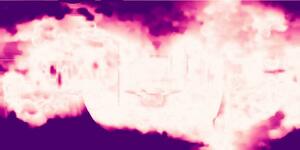}{0.165}{(h) $DIS\_SEPE$}{black} &
		\labeledfig{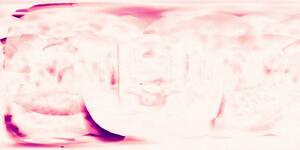}{0.165}{(i) $Our\_SEPE$}{black} &
		\labeledfig{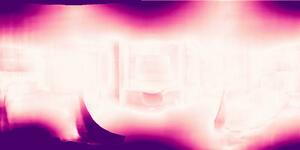}{0.165}{(j) $PWCNet\_SEPE$}{black} &
		\labeledfig{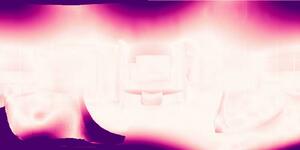}{0.165}{(k) $RAFT\_SEPE$}{black} &
		\labeledfig{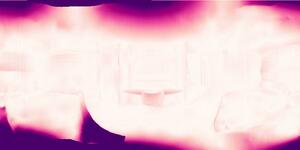}{0.165}{(l) $OmniFlowNet$}{black}
		\\[-0.6em]
		\multicolumn{5}{l}{\footnotesize{\emph{Line}:\emph{office\_1}}}
		\\[-0.6em]
		\labeledfig{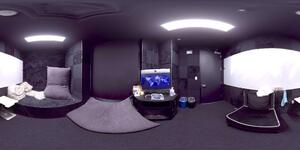}{0.165}{(a) $I_t$}{white} &
		\labeledfig{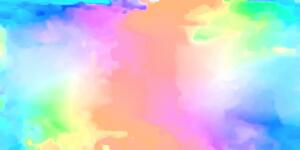}{0.165}{(b) $DIS$}{black} &
		\labeledfig{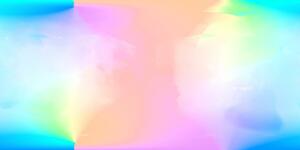}{0.165}{(c) $Our$}{black} &
		\labeledfig{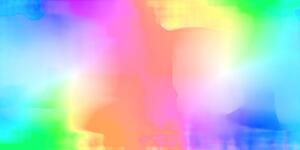}{0.165}{(d) $PWCNet$}{black} &
		\labeledfig{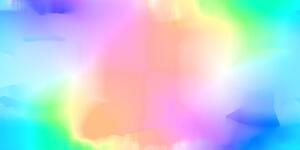}{0.165}{(e) $RAFT$}{black} &
		\labeledfig{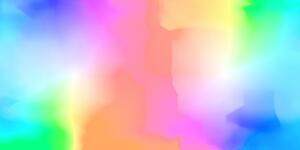}{0.165}{(f) $OmniFlowNet$}{black}
		\\[-1.0em]
		\labeledfig{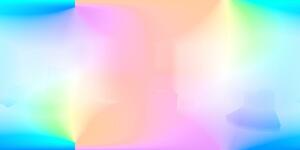}{0.165}{(g) GT}{black} &
		\labeledfig{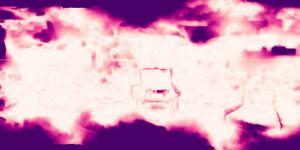}{0.165}{(h) $DIS\_SEPE$}{black} &
		\labeledfig{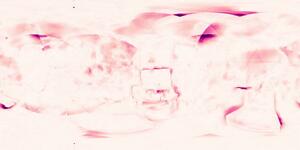}{0.165}{(i) $Our\_SEPE$}{black} &
		\labeledfig{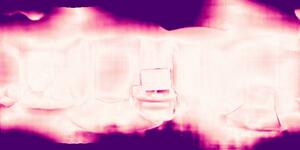}{0.165}{(j) $PWCNet\_SEPE$}{black} &
		\labeledfig{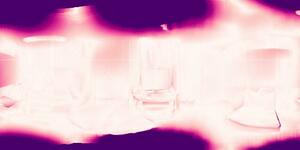}{0.165}{(k) $RAFT\_SEPE$}{black} &
		\labeledfig{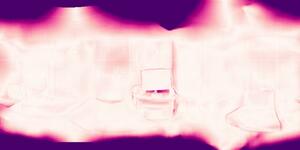}{0.165}{(l) $OmniFlowNet\_SEPE$}{black}
		\\[-0.6em]
		\multicolumn{5}{l}{\footnotesize{\emph{Line}:\emph{room\_0}}}
		\\[-0.6em]
		\labeledfig{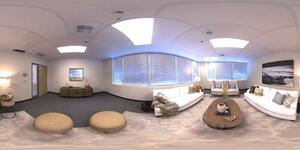}{0.165}{(a) $I_t$}{white} &
		\labeledfig{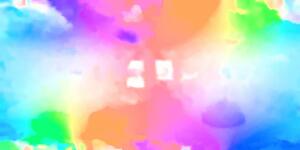}{0.165}{(b) $DIS$}{black} &
		\labeledfig{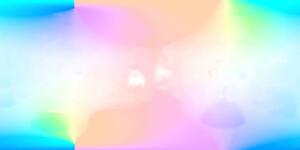}{0.165}{(c) $Our$}{black} &
		\labeledfig{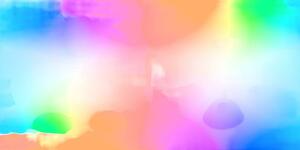}{0.165}{(d) $PWCNet$}{black} &
		\labeledfig{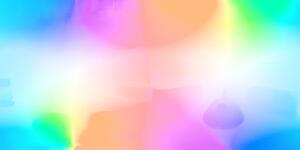}{0.165}{(e) $RAFT$}{black} &
		\labeledfig{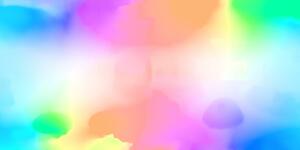}{0.165}{(f) $OmniFlowNet$}{black}
		\\[-1.0em]
		\labeledfig{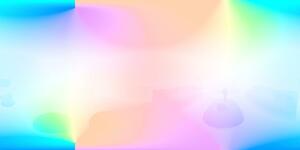}{0.165}{(g) GT}{black} &
		\labeledfig{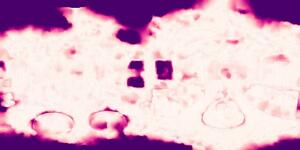}{0.165}{(h) $DIS\_SEPE$}{black} &
		\labeledfig{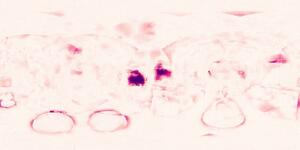}{0.165}{(i) $Our\_SEPE$}{black} &
		\labeledfig{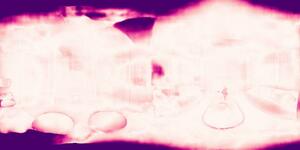}{0.165}{(j) $PWCNet\_SEPE$}{black} &
		\labeledfig{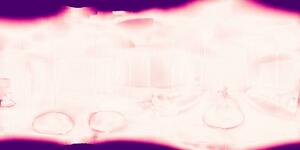}{0.165}{(k) $RAFT\_SEPE$}{black} &
		\labeledfig{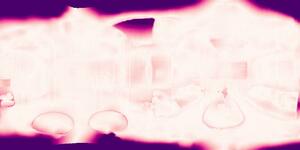}{0.165}{(l) $OmniFlowNet\_SEPE$}{black}
		\\[-0.6em]
		\multicolumn{5}{l}{\footnotesize{\emph{Line}:\emph{room\_1}}}
		\\[-0.6em]
		\labeledfig{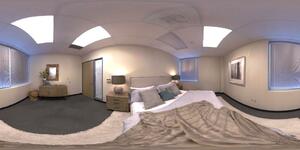}{0.165}{(a) $I_t$}{black} &
		\labeledfig{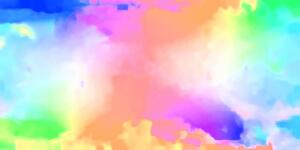}{0.165}{(b) $DIS$}{black} &
		\labeledfig{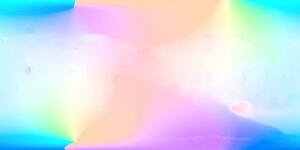}{0.165}{(c) $Our$}{black} &
		\labeledfig{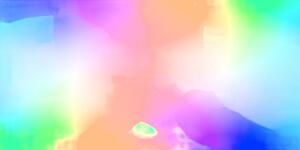}{0.165}{(d) $PWCNet$}{black} &
		\labeledfig{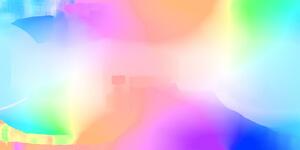}{0.165}{(e) $RAFT$}{black} &
		\labeledfig{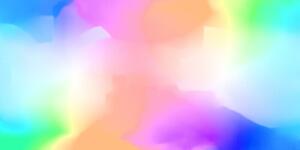}{0.165}{(f) $OmniFlowNet$}{black}
		\\[-1.0em]
		\labeledfig{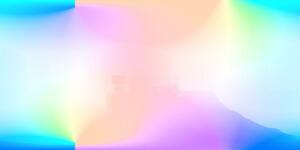}{0.165}{(g) GT}{black} &
		\labeledfig{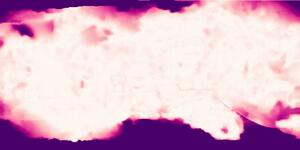}{0.165}{(h) $DIS\_SEPE$}{black} &
		\labeledfig{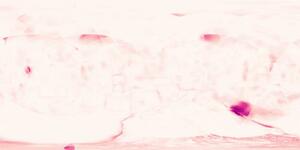}{0.165}{(i) $Our\_SEPE$}{black} &
		\labeledfig{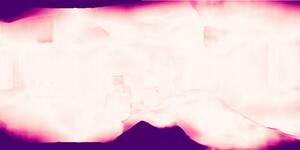}{0.165}{(j) $PWCNet\_SEPE$}{black} &
		\labeledfig{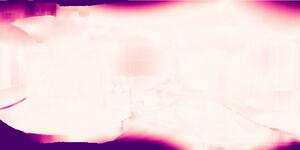}{0.165}{(k) $RAFT\_SEPE$}{black} &
		\labeledfig{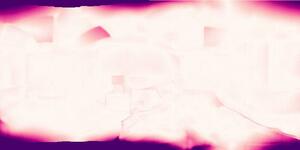}{0.165}{(l) $OmniFlowNet$}{black}
\end{tabular}
\caption{\label{fig:sup:compflowline}%
	Estimated 360° optical flow and error heatmaps on the Replica 360° dataset (\emph{Line}).}
\end{figure*}

\begin{figure*}[hbt!]
	\centering
	\begin{tabular}{@{}c*{6}{@{\hspace{-9pt}}c}@{}}
		\\[-0.5em]
		\multicolumn{5}{l}{\footnotesize{\emph{Line}:\emph{apartment\_0}}}
		\\[-0.5em]
		\labeledfig{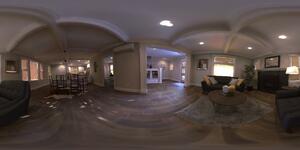}{0.165}{(a) $I_t$}{white} &
		\labeledfig{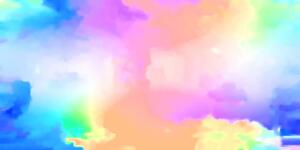}{0.165}{(b) $DIS$}{black} &
		\labeledfig{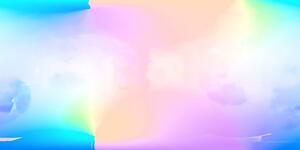}{0.165}{(c) $Our$}{black} &
		\labeledfig{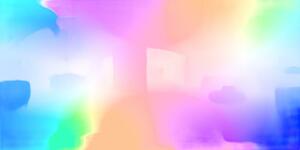}{0.165}{(d) $PWCNet$}{black} &
		\labeledfig{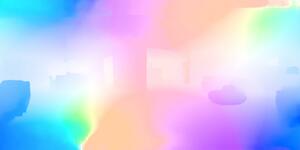}{0.165}{(e) $RAFT$}{black} &
		\labeledfig{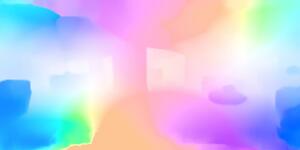}{0.165}{(f) $OmniFlowNet$}{black}
		\\[-1.0em]
		\labeledfig{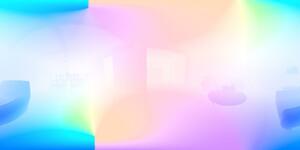}{0.165}{(g) GT}{black} &
		\labeledfig{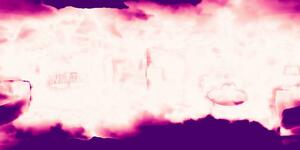}{0.165}{(h) $DIS\_SEPE$}{black} &
		\labeledfig{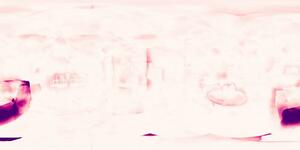}{0.165}{(i) $Our\_SEPE$}{black} &
		\labeledfig{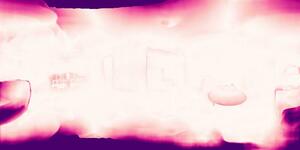}{0.165}{(j) $PWCNet\_SEPE$}{black} &
		\labeledfig{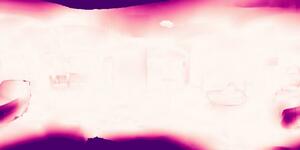}{0.165}{(k) $RAFT\_SEPE$}{black} &
		\labeledfig{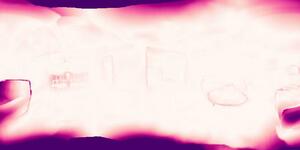}{0.165}{(l) $OmniFlowNet\_SEPE$}{black}
		\\[-0.6em]
		\multicolumn{5}{l}{\footnotesize{\emph{Line}:\emph{apartment\_1}}}
		\\[-0.5em]
		\labeledfig{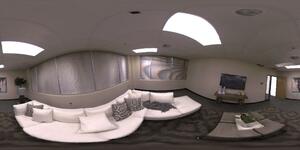}{0.165}{(a) $I_t$}{white} &
		\labeledfig{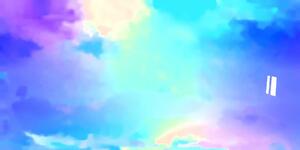}{0.165}{(b) $DIS$}{black} &
		\labeledfig{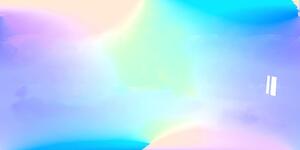}{0.165}{(c) $Our$}{black} &
		\labeledfig{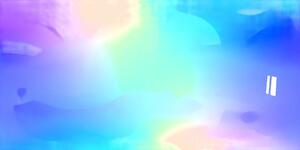}{0.165}{(d) $PWCNet$}{black} &
		\labeledfig{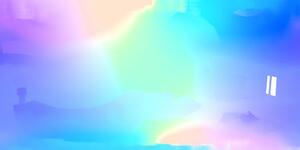}{0.165}{(e) $RAFT$}{black} &
		\labeledfig{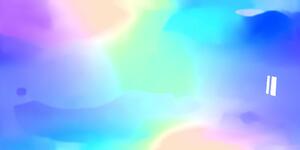}{0.165}{(f) $OmniFlowNet$}{black}
		\\[-1.0em]
		\labeledfig{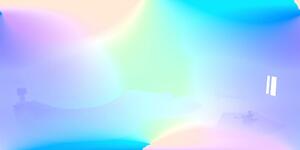}{0.165}{(g) GT}{black} &
		\labeledfig{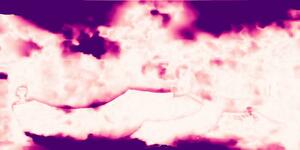}{0.165}{(h) $DIS\_SEPE$}{black} &
		\labeledfig{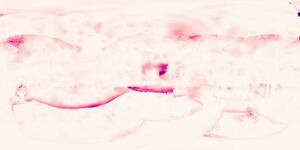}{0.165}{(i) $Our\_SEPE$}{black} &
		\labeledfig{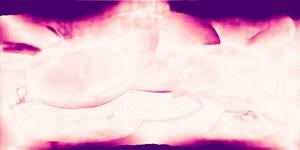}{0.165}{(j) $PWCNet\_SEPE$}{black} &
		\labeledfig{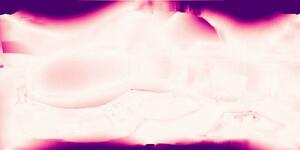}{0.165}{(k) $RAFT\_SEPE$}{black} &
		\labeledfig{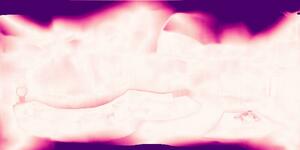}{0.165}{(l) $OmniFlowNet\_SEPE$}{black}
		\\[-0.6em]
		\multicolumn{5}{l}{\footnotesize{\emph{Line}:\emph{frl\_apartment\_1}}}
		\\[-0.6em]
		\labeledfig{images/of_error/frl_apartment_0_rand/frl_apartment_0_rand_1k_0_0004_rgb_pano.jpg}{0.165}{(a) $I_t$}{black} &
		\labeledfig{images/of_error/frl_apartment_0_rand/frl_apartment_0_rand_1k_0_dis_0004_opticalflow_backward_pano.flo_flow_vis.jpg}{0.165}{(b) $DIS$}{black} &
		\labeledfig{images/of_error/frl_apartment_0_rand/frl_apartment_0_rand_1k_0_our_0.4_0004_opticalflow_backward_pano.flo_flow_vis.jpg}{0.165}{(c) $Our$}{black} &
		\labeledfig{images/of_error/frl_apartment_0_rand/frl_apartment_0_rand_1k_0_pwcnet_0004_opticalflow_backward_pano.flo_flow_vis.jpg}{0.165}{(d) $PWCNet$}{black} &
		\labeledfig{images/of_error/frl_apartment_0_rand/frl_apartment_0_rand_1k_0_raft_0004_opticalflow_backward_pano.flo_flow_vis.jpg}{0.165}{(e) $RAFT$}{black} &
		\labeledfig{images/of_error/frl_apartment_0_rand/frl_apartment_0_rand_1k_0_omniflownet_0004_opticalflow_backward_pano.flo_flow_vis.jpg}{0.165}{(f) $OmniFlowNet$}{black}
		\\[-1.0em]
		\labeledfig{images/of_error/frl_apartment_0_rand/frl_apartment_0_rand_1k_0_gt_0004_opticalflow_backward_pano.flo_flow_vis.jpg}{0.165}{(g) GT}{black} &
		\labeledfig{images/of_error/frl_apartment_0_rand/frl_apartment_0_rand_1k_0_dis_0004_opticalflow_backward_pano.flo_error_vis.jpg}{0.165}{(h) $DIS\_SEPE$}{black} &
		\labeledfig{images/of_error/frl_apartment_0_rand/frl_apartment_0_rand_1k_0_our_0.4_0004_opticalflow_backward_pano.flo_error_vis.jpg}{0.165}{(i) $Our\_SEPE$}{black} &
		\labeledfig{images/of_error/frl_apartment_0_rand/frl_apartment_0_rand_1k_0_pwcnet_0004_opticalflow_backward_pano.flo_error_vis.jpg}{0.165}{(j) $PWCNet\_SEPE$}{black} &
		\labeledfig{images/of_error/frl_apartment_0_rand/frl_apartment_0_rand_1k_0_raft_0004_opticalflow_backward_pano.flo_error_vis.jpg}{0.165}{(k) $RAFT\_SEPE$}{black} &
		\labeledfig{images/of_error/frl_apartment_0_rand/frl_apartment_0_rand_1k_0_omniflownet_0004_opticalflow_backward_pano.flo_error_vis.jpg}{0.165}{(l) $OmniFlowNet\_SEPE$}{black}
		\\[-0.6em]
		\multicolumn{5}{l}{\footnotesize{\emph{Line}:\emph{hotel\_0}}}
		\\[-0.6em]
		\labeledfig{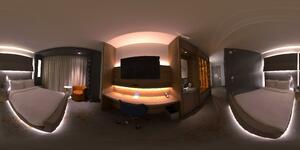}{0.165}{(a) $I_t$}{white} &
		\labeledfig{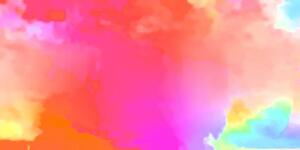}{0.165}{(b) $DIS$}{black} &
		\labeledfig{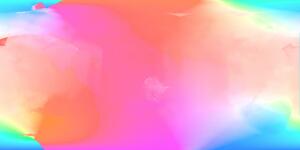}{0.165}{(c) $Our$}{black} &
		\labeledfig{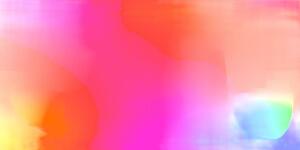}{0.165}{(d) $PWCNet$}{black} &
		\labeledfig{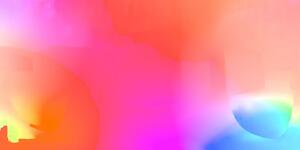}{0.165}{(e) $RAFT$}{black} &
		\labeledfig{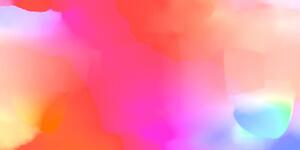}{0.165}{(f) $OmniFlowNet$}{black}
		\\[-1.0em]
		\labeledfig{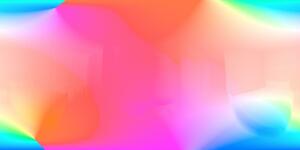}{0.165}{(g) GT}{black} &
		\labeledfig{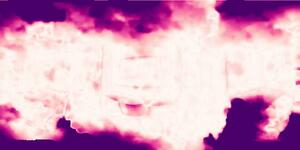}{0.165}{(h) $DIS\_SEPE$}{black} &
		\labeledfig{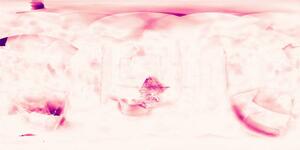}{0.165}{(i) $Our\_SEPE$}{black} &
		\labeledfig{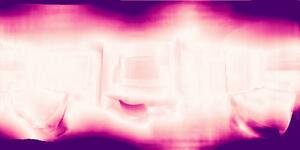}{0.165}{(j) $PWCNet\_SEPE$}{black} &
		\labeledfig{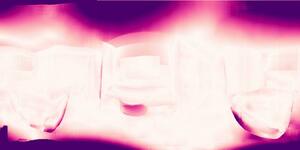}{0.165}{(k) $RAFT\_SEPE$}{black} &
		\labeledfig{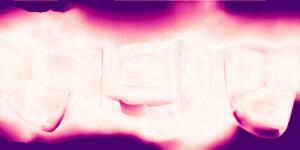}{0.165}{(l) $OmniFlowNet\_SEPE$}{black}
		\\[-0.6em]
		\multicolumn{5}{l}{\footnotesize{\emph{Line}:\emph{office\_1}}}
		\\[-0.6em]
		\labeledfig{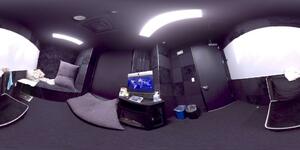}{0.165}{(a) $I_t$}{white} &
		\labeledfig{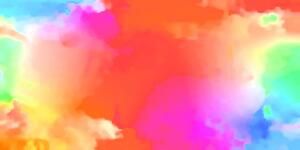}{0.165}{(b) $DIS$}{black} &
		\labeledfig{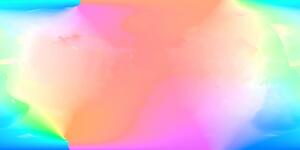}{0.165}{(c) $Our$}{black} &
		\labeledfig{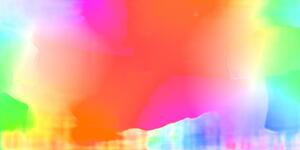}{0.165}{(d) $PWCNet$}{black} &
		\labeledfig{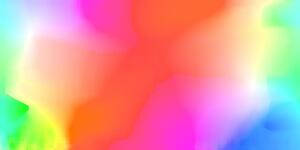}{0.165}{(e) $RAFT$}{black} &
		\labeledfig{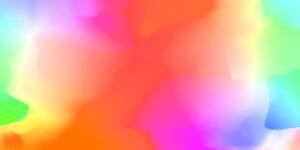}{0.165}{(f) $OmniFlowNet$}{black}
		\\[-1.0em]
		\labeledfig{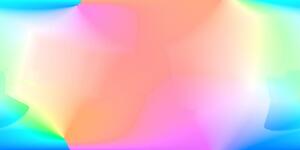}{0.165}{(g) GT}{black} &
		\labeledfig{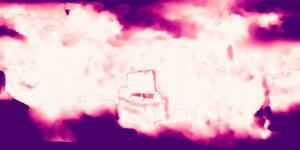}{0.165}{(h) $DIS\_SEPE$}{black} &
		\labeledfig{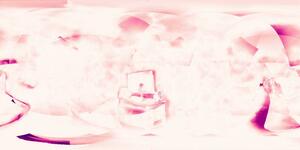}{0.165}{(i) $Our\_SEPE$}{black} &
		\labeledfig{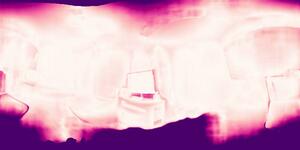}{0.165}{(j) $PWCNet\_SEPE$}{black} &
		\labeledfig{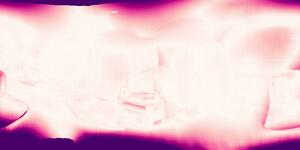}{0.165}{(k) $RAFT\_SEPE$}{black} &
		\labeledfig{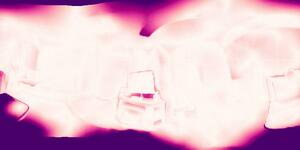}{0.165}{(l) $OmniFlowNet\_SEPE$}{black}
		\\[-0.6em]
		\multicolumn{5}{l}{\footnotesize{\emph{Line}:\emph{room\_0}}}
		\\[-0.6em]
		\labeledfig{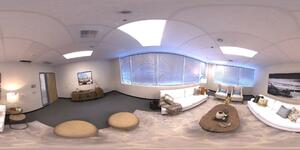}{0.165}{(a) $I_t$}{white} &
		\labeledfig{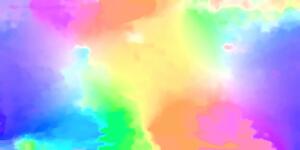}{0.165}{(b) $DIS$}{black} &
		\labeledfig{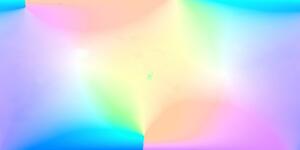}{0.165}{(c) $Our$}{black} &
		\labeledfig{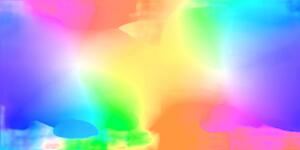}{0.165}{(d) $PWCNet$}{black} &
		\labeledfig{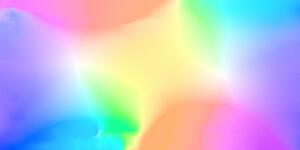}{0.165}{(e) $RAFT$}{black} &
		\labeledfig{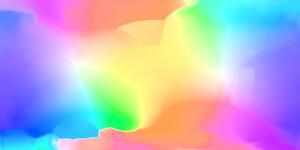}{0.165}{(f) $OmniFlowNet$}{black}
		\\[-1.0em]
		\labeledfig{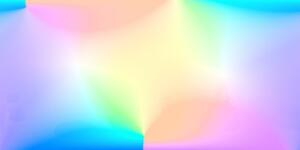}{0.165}{(g) GT}{black} &
		\labeledfig{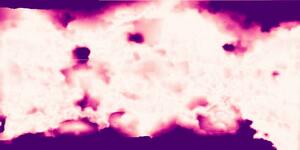}{0.165}{(h) $DIS\_SEPE$}{black} &
		\labeledfig{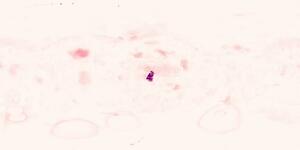}{0.165}{(i) $Our\_SEPE$}{black} &
		\labeledfig{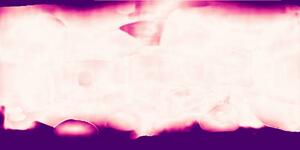}{0.165}{(j) $PWCNet\_SEPE$}{black} &
		\labeledfig{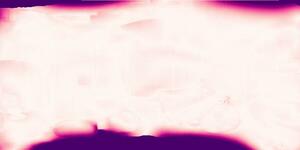}{0.165}{(k) $RAFT\_SEPE$}{black} &
		\labeledfig{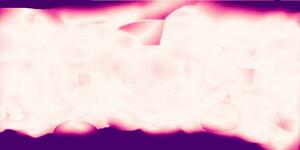}{0.165}{(l) $OmniFlowNet\_SEPE$}{black}
		\\[-0.6em]
		\multicolumn{5}{l}{\footnotesize{\emph{Line}:\emph{room\_1}}}
		\\[-0.6em]
		\labeledfig{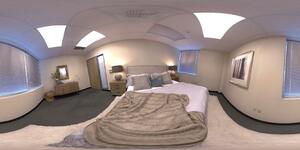}{0.165}{(a) $I_t$}{white} &
		\labeledfig{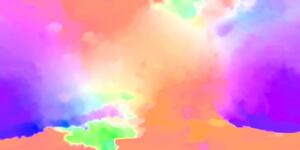}{0.165}{(b) $DIS$}{black} &
		\labeledfig{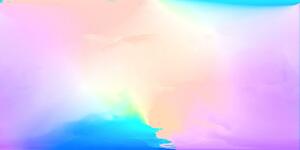}{0.165}{(c) $Our$}{black} &
		\labeledfig{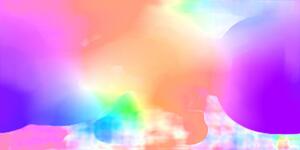}{0.165}{(d) $PWCNet$}{black} &
		\labeledfig{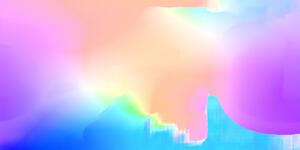}{0.165}{(e) $RAFT$}{black} &
		\labeledfig{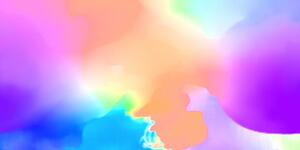}{0.165}{(f) $OmniFlowNet$}{black}
		\\[-1.0em]
		\labeledfig{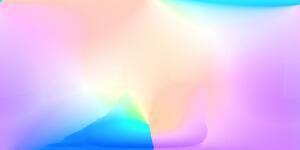}{0.165}{(g) GT}{black} &
		\labeledfig{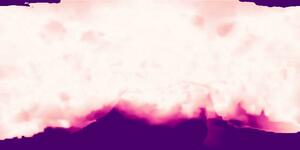}{0.165}{(h) $DIS\_SEPE$}{black} &
		\labeledfig{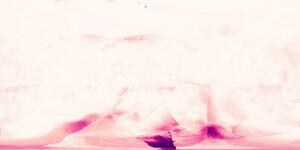}{0.165}{(i) $Our\_SEPE$}{black} &
		\labeledfig{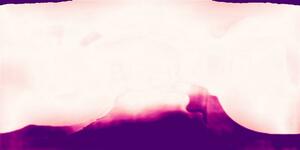}{0.165}{(j) $PWCNet\_SEPE$}{black} &
		\labeledfig{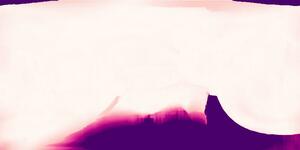}{0.165}{(k) $RAFT\_SEPE$}{black} &
		\labeledfig{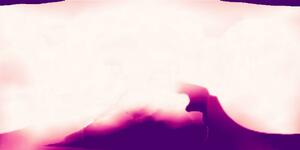}{0.165}{(l) $OmniFlowNet\_SEPE$}{black}
\end{tabular}
\caption{\label{fig:sup:compflowrandom}%
	Estimated 360° optical flow and error heatmaps on the Replica 360° dataset (\emph{Line}).}
\end{figure*}

\cref{fig:sup:compflowcircle}, \cref{fig:sup:compflowline} and \cref{fig:sup:compflowrandom} show additional results and comparisons on the Replica 360° dataset, for circula, linear and random camera motions, respectively.
The our method's SEPE heatmaps show consistently smaller errors near the poles (top and bottom image edges).

\paragraph{Qualitative Evaluation.}

\begin{figure*}[hbt!]
	\centering
	\begin{tabular}{@{}c*{6}{@{\hspace{-9pt}}c}@{}}
		\\[-0.8em]
		\multicolumn{5}{l}{\footnotesize{\emph{Ballintoy}}}
		\\[-0.6em]
		\labeledfig{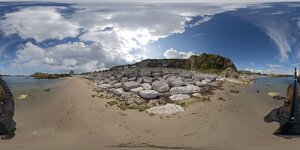}{0.165}{(a) $I_t$}{black} &
		\labeledfig{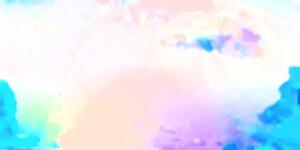}{0.165}{(b) DIS}{black} &
		\labeledfig{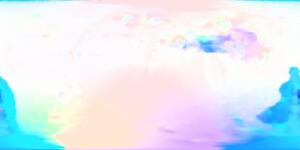}{0.165}{(c) Ours}{black} &
		\labeledfig{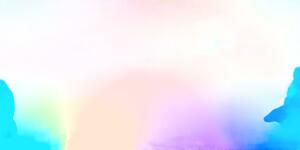}{0.165}{(d) PWC-Net}{black} &
		\labeledfig{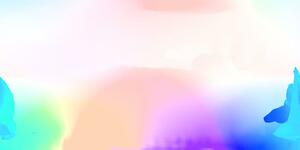}{0.165}{(e) RAFT}{black} &
		\labeledfig{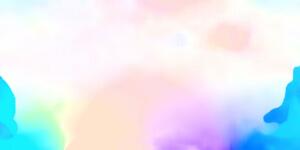}{0.165}{(f) OmniFlowNet}{black}
		\\[-1.0em]
		\labeledfig{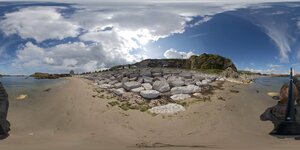}{0.165}{(g) $I_{t+1}$}{black} &
		\labeledfig{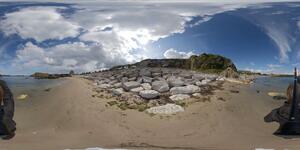}{0.165}{(h) DIS (Warp)}{black} &
		\labeledfig{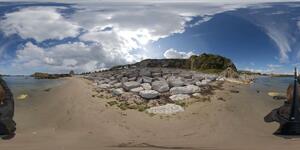}{0.165}{(i) Ours (Warp)}{black} &
		\labeledfig{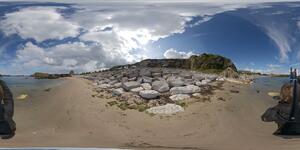}{0.165}{(j) PWC-Net (Warp)}{black} &
		\labeledfig{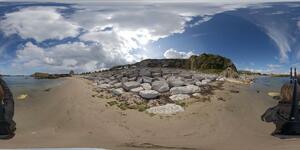}{0.165}{(k) RAFT (Warp)}{black} &
		\labeledfig{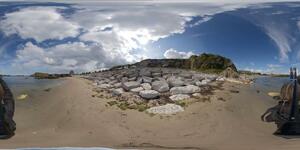}{0.165}{(l) OmniFlowNet (Warp)}{black}
		\\[-1.0em]
		&
		\labeledfig{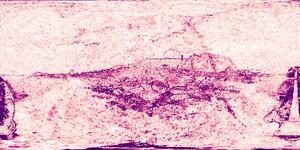}{0.165}{(m) DIS (Warp)}{black} &
		\labeledfig{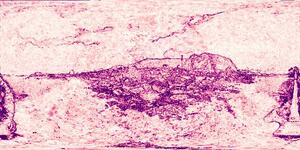}{0.165}{(n) Ours (Warp)}{black} &
		\labeledfig{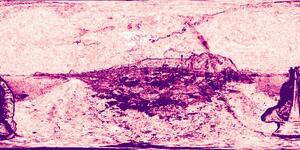}{0.165}{(o) PWC-Net (Warp)}{black} &
		\labeledfig{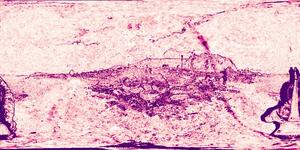}{0.165}{(p) RAFT (Warp)}{black} &
		\labeledfig{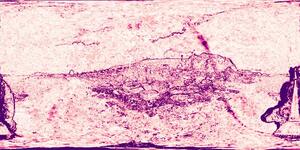}{0.165}{(q) OmniFlowNet (Warp)}{black}
		\\[-0.8em]
		\multicolumn{5}{l}{\footnotesize{\emph{Ship}}}
		\\[-0.6em]
		\labeledfig{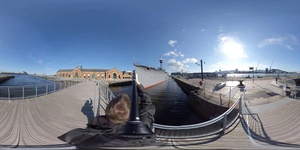}{0.165}{(a) $I_t$}{black} &
		\labeledfig{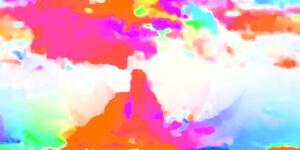}{0.165}{(b) DIS}{black} &
		\labeledfig{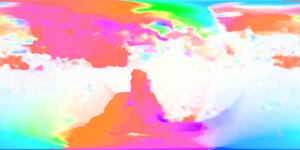}{0.165}{(c) Ours}{black} &
		\labeledfig{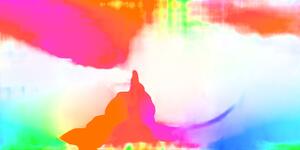}{0.165}{(d) PWC-Net}{black} &
		\labeledfig{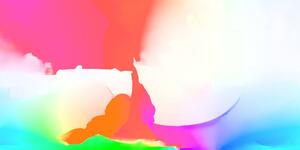}{0.165}{(e) RAFT}{black} &
		\labeledfig{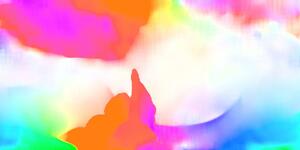}{0.165}{(f) OmniFlowNet}{black}
		\\[-1.0em]
		\labeledfig{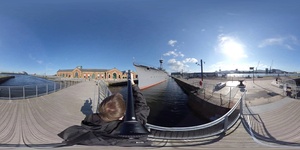}{0.165}{(g) $I_{t+1}$}{black} &
		\labeledfig{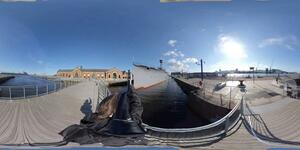}{0.165}{(h) DIS (Warp)}{black} &
		\labeledfig{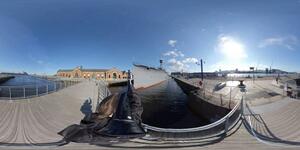}{0.165}{(i) Ours (Warp)}{black} &
		\labeledfig{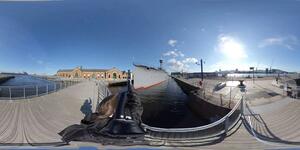}{0.165}{(j) PWC-Net (Warp)}{black} &
		\labeledfig{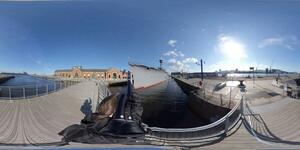}{0.165}{(k) RAFT (Warp)}{black} &
		\labeledfig{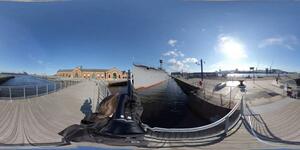}{0.165}{(l) OmniFlowNet (Warp)}{black}
		\\[-1.0em]
		&
		\labeledfig{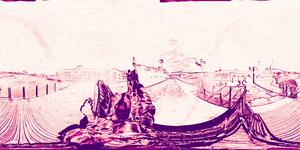}{0.165}{(m) DIS (Warp)}{black} &
		\labeledfig{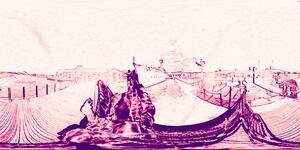}{0.165}{(n) Ours (Warp)}{black} &
		\labeledfig{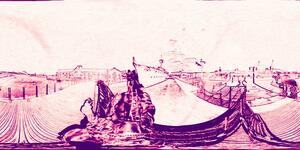}{0.165}{(o) PWC-Net (Warp)}{black} &
		\labeledfig{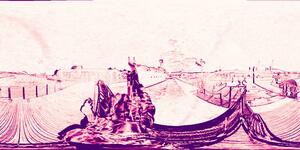}{0.165}{(p) RAFT (Warp)}{black} &
		\labeledfig{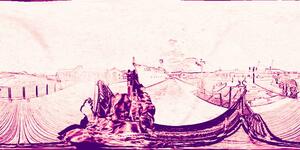}{0.165}{(q) OmniFlowNet (Warp)}{black}
		\\[-0.8em]
		\multicolumn{5}{l}{\footnotesize{\emph{Temple3}}}
		\\[-0.6em]
		\labeledfig{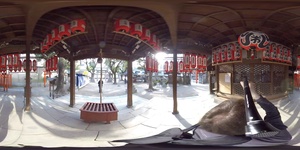}{0.165}{(a) $I_t$}{white} &
		\labeledfig{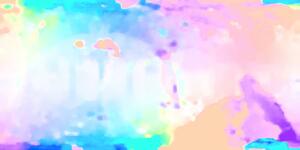}{0.165}{(b) DIS}{black} &
		\labeledfig{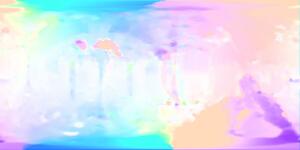}{0.165}{(c) Ours}{black} &
		\labeledfig{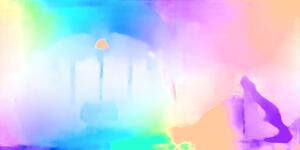}{0.165}{(d) PWC-Net}{black} &
		\labeledfig{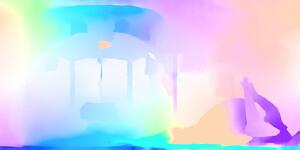}{0.165}{(e) RAFT}{black} &
		\labeledfig{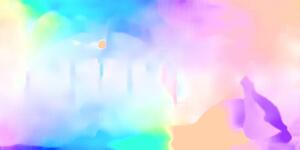}{0.165}{(f) OmniFlowNet}{black}
		\\[-1.0em]
		\labeledfig{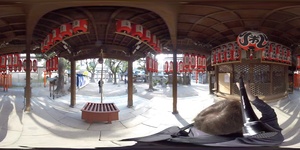}{0.165}{(g) $I_{t+1}$}{white} &
		\labeledfig{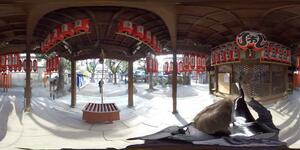}{0.165}{(h) DIS (Warp)}{white} &
		\labeledfig{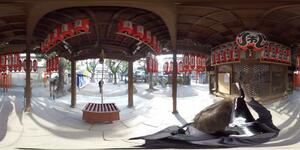}{0.165}{(i) Ours (Warp)}{white} &
		\labeledfig{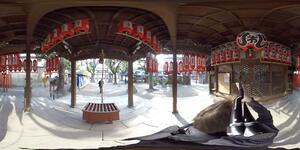}{0.165}{(j) PWC-Net (Warp)}{white} &
		\labeledfig{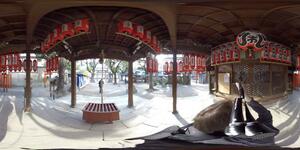}{0.165}{(k) RAFT (Warp)}{white} &
		\labeledfig{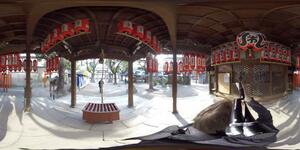}{0.165}{(l) OmniFlowNet (Warp)}{white}
		\\[-1.0em]
		&
		\labeledfig{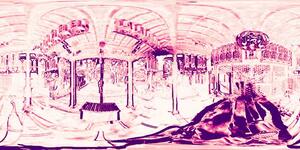}{0.165}{(m) DIS (Warp)}{black} &
		\labeledfig{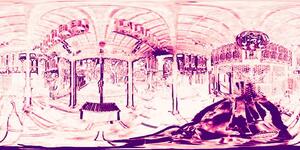}{0.165}{(n) Ours (Warp)}{black} &
		\labeledfig{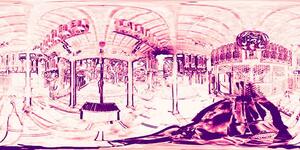}{0.165}{(o) PWC-Net (Warp)}{black} &
		\labeledfig{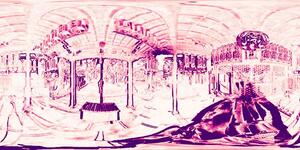}{0.165}{(p) RAFT (Warp)}{black} &
		\labeledfig{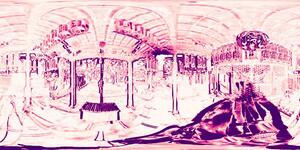}{0.165}{(q) OmniFlowNet (Warp)}{black} 	
		\\[-0.8em]
		\multicolumn{5}{l}{\footnotesize{\emph{SecretGarden1}}}
		\\[-0.6em]
		\labeledfig{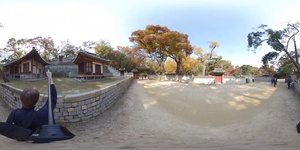}{0.165}{(a) $I_t$}{black} &
		\labeledfig{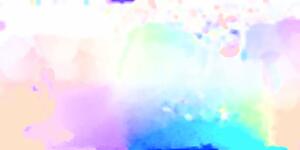}{0.165}{(b) DIS}{black} &
		\labeledfig{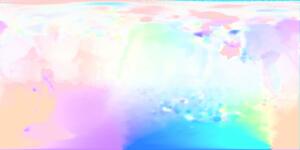}{0.165}{(c) Ours}{black} &
		\labeledfig{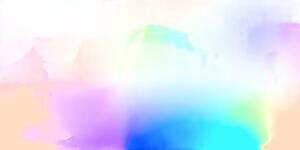}{0.165}{(d) PWC-Net}{black} &
		\labeledfig{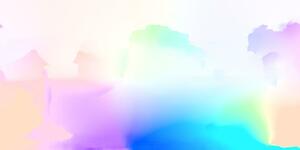}{0.165}{(e) RAFT}{black} &
		\labeledfig{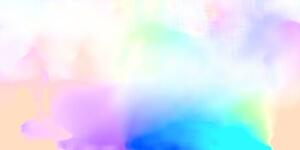}{0.165}{(f) OmniFlowNet}{black}
		\\[-1.0em]
		\labeledfig{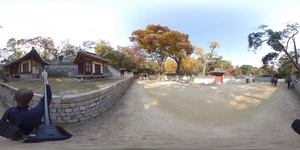}{0.165}{(g) $I_{t+1}$}{black} &
		\labeledfig{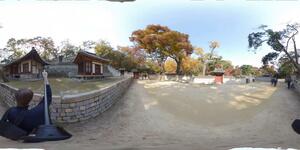}{0.165}{(h) DIS (Warp)}{black} &
		\labeledfig{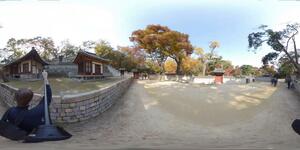}{0.165}{(i) Ours (Warp)}{black} &
		\labeledfig{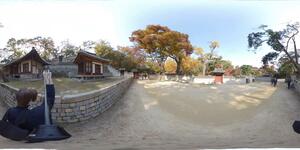}{0.165}{(j) PWC-Net (Warp)}{black} &
		\labeledfig{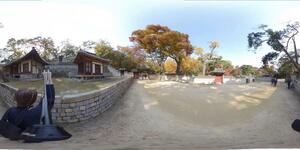}{0.165}{(k) RAFT (Warp)}{black} &
		\labeledfig{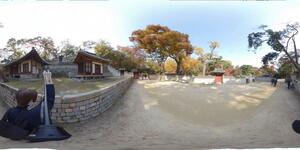}{0.165}{(l) OmniFlowNet (Warp)}{black}
		\\[-1.0em]
		&
		\labeledfig{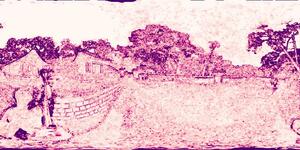}{0.165}{(m) DIS (Warp)}{black} &
		\labeledfig{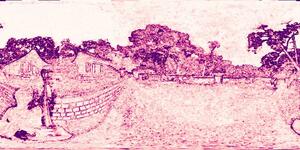}{0.165}{(n) Ours (Warp)}{black} &
		\labeledfig{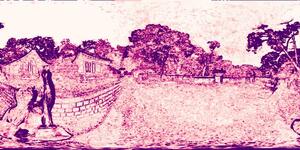}{0.165}{(o) PWC-Net (Warp)}{black} &
		\labeledfig{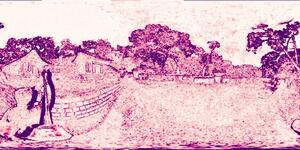}{0.165}{(p) RAFT (Warp)}{black} &
		\labeledfig{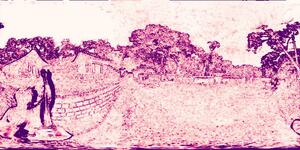}{0.165}{(q) OmniFlowNet (Warp)}{black}	
		\\[-0.8em]
		\multicolumn{5}{l}{\footnotesize{\emph{Wulongting}}}
		\\[-0.6em]
		\labeledfig{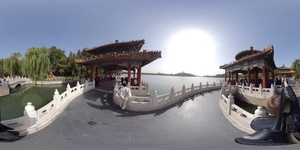}{0.165}{(a) $I_t$}{black} &
		\labeledfig{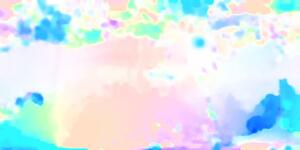}{0.165}{(b) DIS}{black} &
		\labeledfig{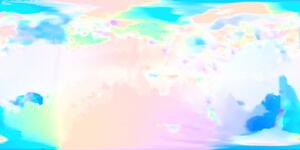}{0.165}{(c) Ours}{black} &
		\labeledfig{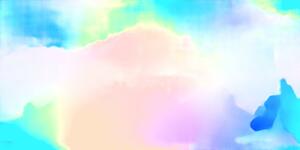}{0.165}{(d) PWC-Net}{black} &
		\labeledfig{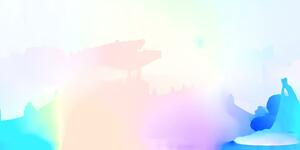}{0.165}{(e) RAFT}{black} &
		\labeledfig{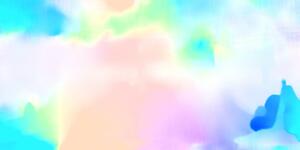}{0.165}{(f) OmniFlowNet}{black}
		\\[-1.0em]
		\labeledfig{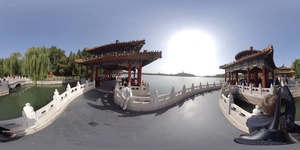}{0.165}{(g) $I_{t+1}$}{black} &
		\labeledfig{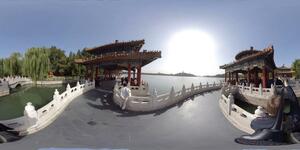}{0.165}{(h) DIS (Warp)}{black} &
		\labeledfig{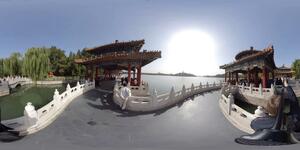}{0.165}{(i) Ours (Warp)}{black} &
		\labeledfig{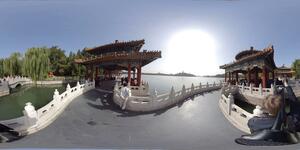}{0.165}{(j) PWC-Net (Warp)}{black} &
		\labeledfig{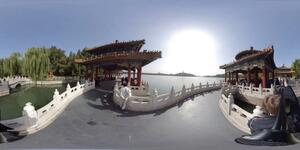}{0.165}{(k) RAFT (Warp)}{black} &
		\labeledfig{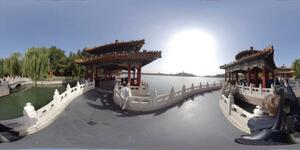}{0.165}{(l) OmniFlowNet (Warp)}{black}
		\\[-1.0em]
		&
		\labeledfig{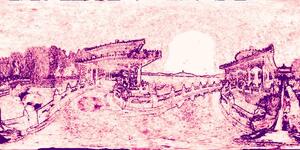}{0.165}{(m) DIS (Warp)}{black} &
		\labeledfig{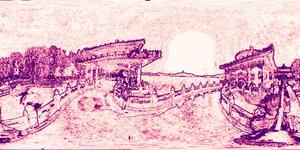}{0.165}{(n) Ours (Warp)}{black} &
		\labeledfig{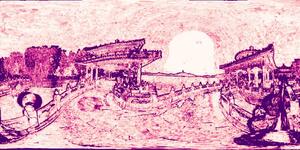}{0.165}{(o) PWC-Net (Warp)}{black} &
		\labeledfig{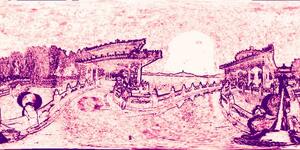}{0.165}{(p) RAFT (Warp)}{black} &
		\labeledfig{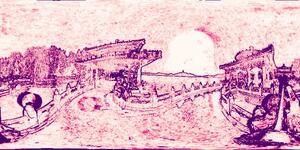}{0.165}{(q) OmniFlowNet (Warp)}{black} 
	\end{tabular}
	\caption{\label{fig:sup:backwardwarp}%
		Backward warping results on several OmniPhotos datasets.}
\end{figure*}

In \cref{fig:sup:backwardwarp}, we show additional interpolation error heatmaps on the OmniPhotos dataset \cite{BerteYLR2020}.
They visualize the interpolation error \cite{BakerSLRBS2011}, i.e. the RGB colour difference between the source image and backward-warped target image.

\end{appendices}

\clearpage
\bibliography{360flow} %

\begin{thebibliography}{51}
\providecommand{\natexlab}[1]{#1}
\providecommand{\url}[1]{\texttt{#1}}
\expandafter\ifx\csname urlstyle\endcsname\relax
  \providecommand{\doi}[1]{doi: #1}\else
  \providecommand{\doi}{doi: \begingroup \urlstyle{rm}\Url}\fi

\bibitem[Aleotti et~al.(2021)Aleotti, Poggi, and Mattoccia]{AleotPM2021}
Filippo Aleotti, Matteo Poggi, and Stefano Mattoccia.
\newblock Learning optical flow from still images.
\newblock In \emph{CVPR}, 2021.

\bibitem[Artizzu et~al.(2020)Artizzu, Zhang, Allibert, and
  Demonceaux]{ArtizZAD2020}
Charles-Olivier Artizzu, Haozhou Zhang, Guillaume Allibert, and Cédric
  Demonceaux.
\newblock {OmniFlowNet}: a perspective neural network adaptation for optical
  flow estimation in omnidirectional images.
\newblock In \emph{ICPR}, 2020.
\newblock \doi{10.1109/ICPR48806.2021.9412745}.

\bibitem[Arun et~al.(1987)Arun, Huang, and Blostein]{ArunHB1987}
K.~S. Arun, T.~S. Huang, and S.~D. Blostein.
\newblock Least-squares fitting of two {3-D} point sets.
\newblock \emph{TPAMI}, 9\penalty0 (5):\penalty0 698--700, 1987.
\newblock \doi{10.1109/TPAMI.1987.4767965}.

\bibitem[Baker et~al.(2011)Baker, Scharstein, Lewis, Roth, Black, and
  Szeliski]{BakerSLRBS2011}
Simon Baker, Daniel Scharstein, J.~Lewis, Stefan Roth, Michael Black, and
  Richard Szeliski.
\newblock A database and evaluation methodology for optical flow.
\newblock \emph{IJCV}, 92\penalty0 (1):\penalty0 1--31, 2011.
\newblock \doi{10.1007/s11263-010-0390-2}.

\bibitem[Bertel et~al.(2020)Bertel, Yuan, Lindroos, and Richardt]{BerteYLR2020}
Tobias Bertel, Mingze Yuan, Reuben Lindroos, and Christian Richardt.
\newblock {OmniPhotos}: Casual 360° {VR} photography.
\newblock \emph{ACM Trans. Graph.}, 39\penalty0 (6):\penalty0 267:1--12, 2020.
\newblock \doi{10.1145/3414685.3417770}.

\bibitem[Bhandari et~al.(2020)Bhandari, Zong, and Yan]{BhandZY2020}
Keshav Bhandari, Ziliang Zong, and Yan Yan.
\newblock Revisiting optical flow estimation in 360 videos.
\newblock In \emph{ICPR}, 2020.
\newblock \doi{10.1109/ICPR48806.2021.9412035}.

\bibitem[Black and Anandan(1991)]{BlackA1991}
Michael~J. Black and P.~Anandan.
\newblock Robust dynamic motion estimation over time.
\newblock In \emph{CVPR}, pages 296--302, 1991.
\newblock \doi{10.1109/CVPR.1991.139705}.

\bibitem[Brox et~al.(2004)Brox, Bruhn, Papenberg, and Weickert]{BroxBPW2004}
Thomas Brox, Andrés Bruhn, Nils Papenberg, and Joachim Weickert.
\newblock High accuracy optical flow estimation based on a theory for warping.
\newblock In \emph{ECCV}, volume 3024, pages 25--36, 2004.
\newblock \doi{10.1007/978-3-540-24673-2_3}.

\bibitem[Cheng et~al.(2018)Cheng, Chao, Dong, Wen, Liu, and
  Sun]{ChengCDWLS2018}
Hsien-Tzu Cheng, Chun-Hung Chao, Jin-Dong Dong, Hao-Kai Wen, Tyng-Luh Liu, and
  Min Sun.
\newblock Cube padding for weakly-supervised saliency prediction in 360°
  videos.
\newblock In \emph{CVPR}, pages 1420--1429, 2018.
\newblock \doi{10.1109/CVPR.2018.00154}.

\bibitem[Coors et~al.(2018)Coors, Condurache, and Geiger]{CoorsCG2018}
Benjamin Coors, Alexandru~Paul Condurache, and Andreas Geiger.
\newblock {SphereNet}: Learning spherical representations for detection and
  classification in omnidirectional images.
\newblock In \emph{ECCV}, pages 518--533, 2018.
\newblock \doi{10.1007/978-3-030-01240-3_32}.

\bibitem[Dosovitskiy et~al.(2015)Dosovitskiy, Fischer, Ilg, Hausser, Hazirbas,
  Golkov, van~der Smagt, Cremers, and Brox]{DosovFIHHGSCB2015}
Alexey Dosovitskiy, Philipp Fischer, Eddy Ilg, Philip Hausser, Caner Hazirbas,
  Vladimir Golkov, Patrick van~der Smagt, Daniel Cremers, and Thomas Brox.
\newblock {FlowNet}: Learning optical flow with convolutional networks.
\newblock In \emph{ICCV}, 2015.
\newblock \doi{10.1109/ICCV.2015.316}.

\bibitem[Eder et~al.(2019)Eder, Moulon, and Guan]{EderMG2019}
Marc Eder, Pierre Moulon, and Li~Guan.
\newblock Pano popups: Indoor {3D} reconstruction with a plane-aware network.
\newblock In \emph{3DV}, pages 76--84, 2019.
\newblock \doi{10.1109/3DV.2019.00018}.

\bibitem[Eder et~al.(2020)Eder, Shvets, Lim, and Frahm]{EderSLF2020}
Marc Eder, Mykhailo Shvets, John Lim, and Jan-Michael Frahm.
\newblock Tangent images for mitigating spherical distortion.
\newblock In \emph{CVPR}, 2020.
\newblock \doi{10.1109/CVPR42600.2020.01244}.

\bibitem[Fernandez-Labrador et~al.(2020)Fernandez-Labrador, Facil, Perez-Yus,
  Demonceaux, Civera, and Guerrero]{FernaFPDCG2020}
Clara Fernandez-Labrador, Jose~M. Facil, Alejandro Perez-Yus, Cédric
  Demonceaux, Javier Civera, and Jose~J. Guerrero.
\newblock Corners for layout: End-to-end layout recovery from 360 images.
\newblock \emph{IEEE Robotics and Automation Letters}, 5\penalty0 (2):\penalty0
  1255--1262, 2020.
\newblock \doi{10.1109/LRA.2020.2967274}.

\bibitem[Horn and Schunck(1981)]{HornS1981}
Berthold K.~P. Horn and Brian~G. Schunck.
\newblock Determining optical flow.
\newblock \emph{Artificial Intelligence}, 17:\penalty0 185--203, 1981.
\newblock \doi{10.1016/0004-3702(81)90024-2}.

\bibitem[Huang et~al.(2017)Huang, Chen, Ceylan, and Jin]{HuangCCJ2017}
Jingwei Huang, Zhili Chen, Duygu Ceylan, and Hailin Jin.
\newblock 6-{DOF VR} videos with a single 360-camera.
\newblock In \emph{IEEE VR}, pages 37--44, 2017.
\newblock \doi{10.1109/VR.2017.7892229}.

\bibitem[Hui et~al.(2018)Hui, Tang, and Change~Loy]{HuiTC2018}
Tak-Wai Hui, Xiaoou Tang, and Chen Change~Loy.
\newblock {LiteFlowNet}: A lightweight convolutional neural network for optical
  flow estimation.
\newblock In \emph{CVPR}, pages 8981--8989, 2018.
\newblock \doi{10.1109/CVPR.2018.00936}.

\bibitem[Ilg et~al.(2017)Ilg, Mayer, Saikia, Keuper, Dosovitskiy, and
  Brox]{IlgMSKDB2017}
Eddy Ilg, Nikolaus Mayer, Tonmoy Saikia, Margret Keuper, Alexey Dosovitskiy,
  and Thomas Brox.
\newblock {FlowNet} 2.0: Evolution of optical flow estimation with deep
  networks.
\newblock In \emph{CVPR}, 2017.
\newblock \doi{10.1109/CVPR.2017.179}.

\bibitem[Im et~al.(2016)Im, Ha, Rameau, Jeon, Choe, and Kweon]{ImHRJCK2016}
Sunghoon Im, Hyowon Ha, François Rameau, Hae-Gon Jeon, Gyeongmin Choe, and
  In~So Kweon.
\newblock All-around depth from small motion with a spherical panoramic camera.
\newblock In \emph{ECCV}, 2016.
\newblock \doi{10.1007/978-3-319-46487-9_10}.

\bibitem[Jiang et~al.(2021)Jiang, Sheng, Zhu, Dong, and Huang]{JiangSZDH2021}
Hualie Jiang, Zhe Sheng, Siyu Zhu, Zilong Dong, and Rui Huang.
\newblock {UniFuse}: Unidirectional fusion for 360° panorama depth estimation.
\newblock \emph{IEEE Robotics and Automation Letters}, 6\penalty0 (2):\penalty0
  1519--1526, 2021.
\newblock \doi{10.1109/LRA.2021.3058957}.

\bibitem[Jin et~al.(2020)Jin, Xu, Zheng, Zhang, Tang, Xu, Yu, and
  Gao]{JinXZZTXYG2020}
Lei Jin, Yanyu Xu, Jia Zheng, Junfei Zhang, Rui Tang, Shugong Xu, Jingyi Yu,
  and Shenghua Gao.
\newblock Geometric structure based and regularized depth estimation from 360
  indoor imagery.
\newblock In \emph{CVPR}, pages 886--895, 2020.
\newblock \doi{10.1109/CVPR42600.2020.00097}.

\bibitem[Kopf(2016)]{Kopf2016}
Johannes Kopf.
\newblock 360° video stabilization.
\newblock \emph{ACM Trans. Graph.}, 35\penalty0 (6):\penalty0 195:1--9, 2016.
\newblock \doi{10.1145/2980179.2982405}.

\bibitem[Kroeger et~al.(2016)Kroeger, Timofte, Dai, and Van~Gool]{KroegTDV2016}
Till Kroeger, Radu Timofte, Dengxin Dai, and Luc Van~Gool.
\newblock Fast optical flow using dense inverse search.
\newblock In \emph{ECCV}, pages 471--488, 2016.
\newblock \doi{10.1007/978-3-319-46493-0_29}.

\bibitem[Lai et~al.(2019)Lai, Xie, Lang, and Laganière]{LaiXLL2019}
Po~Kong Lai, Shuang Xie, Jochen Lang, and Robert Laganière.
\newblock Real-time panoramic depth maps from omni-directional stereo images
  for 6 {DoF} videos in virtual reality.
\newblock In \emph{IEEE VR}, pages 405--412, 2019.
\newblock \doi{10.1109/VR.2019.8798016}.

\bibitem[Lee et~al.(2019)Lee, Jeong, Yun, June, and Yoon]{LeeJYJY2019}
Yeon~Kun Lee, Jaeseok Jeong, Jong~Seob Yun, Cho~Won June, and Kuk-Jin Yoon.
\newblock {SpherePHD}: Applying {CNNs} on a spherical {PolyHeDron}
  representation of 360° images.
\newblock In \emph{CVPR}, pages 9173--9181, 2019.
\newblock \doi{10.1109/CVPR.2019.00940}.

\bibitem[Lucas and Kanade(1981)]{LucasK1981}
Bruce~D. Lucas and Takeo Kanade.
\newblock An iterative image registration technique with an application to
  stereo vision.
\newblock In \emph{Proceedings of the International Joint Conference on
  Artificial Intelligence}, 1981.

\bibitem[Luo et~al.(2019)Luo, Zhang, Su, and Xiang]{LuoZSX2019}
Junren Luo, Wanpeng Zhang, Jiongming Su, and Fengtao Xiang.
\newblock Hexagonal convolutional neural networks for hexagonal grids.
\newblock \emph{IEEE Access}, 7:\penalty0 142738--142749, 2019.
\newblock \doi{10.1109/ACCESS.2019.2944766}.

\bibitem[Matzen et~al.(2017)Matzen, Cohen, Evans, Kopf, and
  Szeliski]{MatzeCEKS2017}
Kevin Matzen, Michael~F. Cohen, Bryce Evans, Johannes Kopf, and Richard
  Szeliski.
\newblock Low-cost 360 stereo photography and video capture.
\newblock \emph{ACM Trans. Graph.}, 36\penalty0 (4):\penalty0 148:1--12, 2017.
\newblock \doi{10.1145/3072959.3073645}.

\bibitem[Menze et~al.(2018)Menze, Heipke, and Geiger]{MenzeHG2018}
Moritz Menze, Christian Heipke, and Andreas Geiger.
\newblock Object scene flow.
\newblock \emph{ISPRS Journal of Photogrammetry and Remote Sensing},
  140:\penalty0 60--76, 2018.
\newblock \doi{10.1016/j.isprsjprs.2017.09.013}.

\bibitem[Shugrina et~al.(2019)Shugrina, Liang, Kar, Li, Singh, Singh, and
  Fidler]{ShugrLKLSSF2019}
Maria Shugrina, Ziheng Liang, Amlan Kar, Jiaman Li, Angad Singh, Karan Singh,
  and Sanja Fidler.
\newblock Creative flow+ dataset.
\newblock In \emph{CVPR}, pages 5384--5393, 2019.
\newblock \doi{10.1109/CVPR.2019.00553}.

\bibitem[Sorkine-Hornung and Rabinovich(2017)]{SorkiR2017}
Olga Sorkine-Hornung and Michael Rabinovich.
\newblock Least-squares rigid motion using {SVD}.
\newblock Note, 2017.

\bibitem[Straub et~al.(2019)Straub, Whelan, Ma, Chen, Wijmans, Green, Engel,
  Mur-Artal, Ren, Verma, Clarkson, Yan, Budge, Yan, Pan, Yon, Zou, Leon,
  Carter, Briales, Gillingham, Mueggler, Pesqueira, Savva, Batra, Strasdat,
  Nardi, Goesele, Lovegrove, and
  Newcombe]{StrauWMCWGEMRVCYBYPYZLCBGMPSBSNGLN2019}
Julian Straub, Thomas Whelan, Lingni Ma, Yufan Chen, Erik Wijmans, Simon Green,
  Jakob~J. Engel, Raul Mur-Artal, Carl Ren, Shobhit Verma, Anton Clarkson,
  Mingfei Yan, Brian Budge, Yajie Yan, Xiaqing Pan, June Yon, Yuyang Zou,
  Kimberly Leon, Nigel Carter, Jesus Briales, Tyler Gillingham, Elias Mueggler,
  Luis Pesqueira, Manolis Savva, Dhruv Batra, Hauke~M. Strasdat, Renzo~De
  Nardi, Michael Goesele, Steven Lovegrove, and Richard Newcombe.
\newblock The {Replica} dataset: A digital replica of indoor spaces.
\newblock arXiv:\href{https://arxiv.org/abs/1906.05797}{1906.05797}, 2019.

\bibitem[Su and Grauman(2019)]{SuG2019}
Yu-Chuan Su and Kristen Grauman.
\newblock Kernel transformer networks for compact spherical convolution.
\newblock In \emph{CVPR}, pages 9442--9451, 2019.
\newblock \doi{10.1109/CVPR.2019.00967}.

\bibitem[Sun et~al.(2021)Sun, Sun, and Chen]{SunSC2021}
Cheng Sun, Min Sun, and Hwann-Tzong Chen.
\newblock {HoHoNet}: 360 indoor holistic understanding with latent horizontal
  features.
\newblock In \emph{CVPR}, 2021.

\bibitem[Sun et~al.(2014)Sun, Roth, and Black]{SunRB2014}
Deqing Sun, Stefan Roth, and Michael~J. Black.
\newblock A quantitative analysis of current practices in optical flow
  estimation and the principles behind them.
\newblock \emph{IJCV}, 106\penalty0 (2):\penalty0 115--137, 2014.
\newblock \doi{10.1007/s11263-013-0644-x}.

\bibitem[Sun et~al.(2018)Sun, Yang, Liu, and Kautz]{SunYLK2018}
Deqing Sun, Xiaodong Yang, Ming-Yu Liu, and Jan Kautz.
\newblock {PWC-Net}: {CNNs} for optical flow using pyramid, warping, and cost
  volume.
\newblock In \emph{CVPR}, pages 8934--8943, 2018.
\newblock \doi{10.1109/CVPR.2018.00931}.

\bibitem[Sun et~al.(2020)Sun, Yang, Liu, and Kautz]{SunYLK2020}
Deqing Sun, Xiaodong Yang, Ming-Yu Liu, and Jan Kautz.
\newblock Models matter, so does training: An empirical study of {CNNs} for
  optical flow estimation.
\newblock \emph{TPAMI}, 42\penalty0 (6):\penalty0 1408--1423, 2020.
\newblock \doi{10.1109/TPAMI.2019.2894353}.

\bibitem[Tateno et~al.(2018)Tateno, Navab, and Tombari]{TatenNT2018}
Keisuke Tateno, Nassir Navab, and Federico Tombari.
\newblock Distortion-aware convolutional filters for dense prediction in
  panoramic images.
\newblock In \emph{ECCV}, pages 732--750, 2018.
\newblock \doi{10.1007/978-3-030-01270-0_43}.

\bibitem[Teed and Deng(2020)]{TeedD2020a}
Zachary Teed and Jia Deng.
\newblock {RAFT}: Recurrent all-pairs field transforms for optical flow.
\newblock In \emph{ECCV}, 2020.
\newblock \doi{10.1007/978-3-030-58536-5_24}.

\bibitem[Tran(2021)]{Tran2021}
Phi~Vu Tran.
\newblock {SSLayout360}: Semi-supervised indoor layout estimation from
  360-degree panorama.
\newblock In \emph{CVPR}, 2021.

\bibitem[Wang et~al.(2018)Wang, Hu, Cheng, Lin, Yang, Shih, Chu, and
  Sun]{WangHCLYSCS2018}
Fu-En Wang, Hou-Ning Hu, Hsien-Tzu Cheng, Juan-Ting Lin, Shang-Ta Yang, Meng-Li
  Shih, Hung-Kuo Chu, and Min Sun.
\newblock Self-supervised learning of depth and camera motion from 360°
  videos.
\newblock In \emph{ACCV}, pages 53--68, 2018.
\newblock \doi{10.1007/978-3-030-20873-8_4}.

\bibitem[Wang et~al.(2020{\natexlab{a}})Wang, Yeh, Sun, Chiu, and
  Tsai]{WangYSCT2020}
Fu-En Wang, Yu-Hsuan Yeh, Min Sun, Wei-Chen Chiu, and Yi-Hsuan Tsai.
\newblock {BiFuse}: Monocular 360 depth estimation via bi-projection fusion.
\newblock In \emph{CVPR}, pages 462--471, 2020{\natexlab{a}}.
\newblock \doi{10.1109/CVPR42600.2020.00054}.

\bibitem[Wang et~al.(2021)Wang, Yeh, Sun, Chiu, and Tsai]{WangYSCT2021}
Fu-En Wang, Yu-Hsuan Yeh, Min Sun, Wei-Chen Chiu, and Yi-Hsuan Tsai.
\newblock {LED$^2$}-net: Monocular 360 layout estimation via differentiable
  depth rendering.
\newblock In \emph{CVPR}, 2021.

\bibitem[Wang et~al.(2020{\natexlab{b}})Wang, Solarte, Tsai, Chiu, and
  Sun]{WangSTCS2020}
Ning-Hsu Wang, Bolivar Solarte, Yi-Hsuan Tsai, Wei-Chen Chiu, and Min Sun.
\newblock {360SD}-net: 360° stereo depth estimation with learnable cost
  volume.
\newblock In \emph{ICRA}, pages 582--588, 2020{\natexlab{b}}.
\newblock \doi{10.1109/ICRA40945.2020.9196975}.

\bibitem[Xu et~al.(2021)Xu, Zheng, Xu, Tang, and Gao]{XuZXTG2021}
Jiale Xu, Jia Zheng, Yanyu Xu, Rui Tang, and Shenghua Gao.
\newblock Layout-guided novel view synthesis from a single indoor panorama.
\newblock In \emph{CVPR}, 2021.

\bibitem[Yang et~al.(2021)Yang, Zhang, Reiß, Hu, and
  Stiefelhagen]{YangZRHS2021}
Kailun Yang, Jiaming Zhang, Simon Reiß, Xinxin Hu, and Rainer Stiefelhagen.
\newblock Capturing omni-range context for omnidirectional segmentation.
\newblock In \emph{CVPR}, 2021.

\bibitem[Zeng et~al.(2020)Zeng, Karaoglu, and Gevers]{ZengKG2020}
Wei Zeng, Sezer Karaoglu, and Theo Gevers.
\newblock Joint {3D} layout and depth prediction from a single indoor panorama
  image.
\newblock In \emph{ECCV}, 2020.
\newblock \doi{10.1007/978-3-030-58517-4_39}.

\bibitem[Zhang et~al.(2019)Zhang, Liwicki, Smith, and Cipolla]{ZhangLSC2019}
Chao Zhang, Stephan Liwicki, William Smith, and Roberto Cipolla.
\newblock Orientation-aware semantic segmentation on icosahedron spheres.
\newblock In \emph{ICCV}, pages 3533--3541, 2019.
\newblock \doi{10.1109/ICCV.2019.00363}.

\bibitem[Zhao et~al.(2020)Zhao, You, Zhang, Liu, Bian, and Tong]{ZhaoYZLBT2020}
Pengyu Zhao, Ansheng You, Yuanxing Zhang, Jiaying Liu, Kaigui Bian, and Yunhai
  Tong.
\newblock Spherical criteria for fast and accurate 360° object detection.
\newblock In \emph{Proceedings of the Conference on Artificial Intelligence
  (AAAI)}, volume~34, pages 12959--12966, 2020.
\newblock \doi{10.1609/aaai.v34i07.6995}.

\bibitem[Zioulis et~al.(2018)Zioulis, Karakottas, Zarpalas, and
  Daras]{ZioulKZD2018}
Nikolaos Zioulis, Antonis Karakottas, Dimitrios Zarpalas, and Petros Daras.
\newblock {OmniDepth}: Dense depth estimation for indoors spherical panoramas.
\newblock In \emph{ECCV}, pages 448--465, 2018.
\newblock \doi{10.1007/978-3-030-01231-1_28}.

\bibitem[Zioulis et~al.(2019)Zioulis, Karakottas, Zarpalas, Alvarez, and
  Daras]{ZioulKZAD2019}
Nikolaos Zioulis, Antonis Karakottas, Dimitrios Zarpalas, Federico Alvarez, and
  Petros Daras.
\newblock Spherical view synthesis for self-supervised 360° depth estimation.
\newblock In \emph{3DV}, pages 690--699, 2019.
\newblock \doi{10.1109/3DV.2019.00081}.

\end{thebibliography}

\end{document}